\documentclass[lettersize,journal]{IEEEtran}
\usepackage{amsmath,amsfonts,amssymb}
\usepackage{algorithmic}
\usepackage{algorithm}
\usepackage{array}
\usepackage[caption=false,font=normalsize,labelfont=sf,textfont=sf]{subfig}
\usepackage{textcomp}
\usepackage{stfloats}
\usepackage{url}
\usepackage{verbatim}
\usepackage{graphicx}
\usepackage{cite}
\usepackage{booktabs}
\usepackage{pifont}
\usepackage{color}
\usepackage[table]{xcolor}
\usepackage{placeins}
\usepackage{tabularx}
\usepackage{ragged2e}
\usepackage{mathrsfs}
\usepackage{hyperref}
\hypersetup{
	colorlinks=true,
	linkcolor=blue,   
	filecolor=magenta,
	urlcolor=cyan,   
	citecolor=blue    
}
\begin{document}

\title{Consistency-Regularized GAN for Few-Shot SAR Target Recognition}

\author{Yikui Zhai,~\IEEEmembership{Senior Member,~IEEE},
	Shikuang Liu,
	Wenlve Zhou,
	Hongsheng Zhang,~\IEEEmembership{Senior Member,~IEEE},
	Zhiheng Zhou,
	Xiaolin Tian
	and C. L. Philip Chen,~\IEEEmembership{Life Fellow,~IEEE}
\thanks{This study was funded by Guangdong Higher Education Innovation and Strengthening School Project (No. 2022ZDZX1032, No. 2023ZDZX1029), Wuyi University Hong Kong and Macao Joint Research and Development Fund (No. 2022WGALH19), Guangdong Jiangmen Science and Technology Research Project (No. 2220002000246, No. 2023760300070008390). This work was supported by the High Performance Computing Center of Wuyi University. (Yikui Zhai, Shikuang Liu and Wenlve Zhou contributed equally to this work) (Corresponding authors: Yikui Zhai and Hongsheng Zhang.)}
\thanks{Yikui Zhai and Shikuang Liu are with the College of Electronics and 	Information Engineering, Wuyi University, Jiangmen, 529020, China (e-mail: yikuizhai@163.com; eeshikuangliu@163.com).\par
	Wenlve Zhou and Zhiheng Zhou are with the College of Electronic and Information Engineering, South China University of Technology, Guangzhou 510641, Guangdong, China, and also with Key Laboratory of Big Data and Intelligent Robot, Ministry of Education, South China University of Technology, Guangzhou 510641, China (email: wenlvezhou@163.com; zhouzh@scut.edu.cn). \par
	Hongsheng Zhang is with the Department of Geography, The University of Hong Kong, Hong Kong, China (e-mail: zhanghs@hku.hk).\par
	Xiaolin Tian is with the Faculty of Innovation Engineering, Macau University of Science and Technology, Macao, China (e-mail: xltian@must.edu.mo).\par
	C. L. Philip Chen is with the Faculty of Computer Science and Engineering,
	South China University of Technology, Guangzhou 510006, China (e-mail:
	philip.chen@ieee.org).
	
}}

% The paper headers
\markboth{Journal of \LaTeX\ Class Files,~Vol.~14, No.~8, August~2021}%
{Shell \MakeLowercase{\textit{et al.}}: A Sample Article Using IEEEtran.cls for IEEE Journals}
\IEEEpubidadjcol

\maketitle
\definecolor{light-gray}{gray}{0.92}
\begin{abstract}
Few-shot recognition in synthetic aperture radar (SAR) imagery remains a critical bottleneck for real-world applications due to extreme data scarcity. A promising strategy involves synthesizing a large dataset with a generative adversarial network (GAN), pre-training a model via self-supervised learning (SSL), and then fine-tuning on the few labeled samples. However, this approach faces a fundamental paradox: conventional GANs themselves require abundant data for stable training, contradicting the premise of few-shot learning.
To resolve this, we propose the consistency-regularized generative adversarial network (Cr-GAN), a novel framework designed to synthesize diverse, high-fidelity samples even when trained under these severe data limitations. Cr-GAN introduces a dual-branch discriminator that decouples adversarial training from representation learning. This architecture enables a channel-wise feature interpolation strategy to create novel latent features, complemented by a dual-domain cycle consistency mechanism that ensures semantic integrity. Our Cr-GAN framework is adaptable to various GAN architectures, and its synthesized data effectively boosts multiple SSL algorithms. Extensive experiments on the MSTAR and SRSDD datasets validate our approach, with Cr-GAN achieving a highly competitive accuracy of 71.21\% and 51.64\%, respectively, in the 8-shot setting, significantly outperforming leading baselines, while requiring only $\sim$5\% of the parameters of state-of-the-art diffusion models. Code is available at: \textit{https://github.com/yikuizhai/Cr-GAN}.
\\
\\
\end{abstract}

\vspace{14pt}
\section{Introduction}
\IEEEPARstart{S}{ar} target recognition has become a pivotal task in a wide range of applications such as land cover mapping, urban planning, and disaster management. However, one of the most significant challenges faced in remote sensing (RS) tasks is the scarcity of labeled data, as acquiring labeled RS images requires considerable resources and expertise. This limitation often results in the underperformance of machine learning models\cite{zhu2017deep}, as they struggle to generalize when trained on small labeled datasets. To address this challenge, there is a growing need for methods that can effectively leverage limited labeled data\cite{tgrs1}, while augmenting the training process with synthetic, high-quality data\cite{tgrs2}.\par

How to use limited data for neural network training is a long-standing problem. Most mainstream approaches fall into one of two classes: SSL \cite{tgrs3,tgrs4,tgrs5,WCL,dca,cgid,LDCL} and meta-learning\cite{ref5,tgrs1,tgrs7,TANG1,asbi}. Pre-training with quantities of unlabeled data and fine-tuning with a few labeled ones is the core of the former. The follow-up years witnesses its great progress \cite{ref1,ref2,ref3,ref4}. Meta-learning aims to quickly adapt to new tasks through few-shot samples and the key is the prior knowledge from the base classes. These methods are the cornerstone techniques for numerous vision applications, such as object detection \cite{ref6,ref7}, semantic segmentation \cite{ref8,ref9} and visual tracking \cite{ref10,ref11}. The success of these methods depends on auxiliary knowledge (e.g., source domain knowledge, base classes, or unlabeled data). However, additional data is still scarce in various scenarios, e.g. RS image with military target, defective material etc. This makes it challenging for these technologies to be widely applied in real world.\par

GANs \cite{ref14,shi2022legan,chen2025data,lin2025hyperking,zhong2024shbgan,goyes2024gan,zhang2024stugan,ge2024waveletgan,li2024transfer,wang2024gpr,xie2024gan,zhu2023qisgan}  seem to have the potential to solve this problem, which is able to synthesize high-fidelity images. Unfortunately, the GAN methods still heavily rely on a large amount of diverse training data. To deal with this shortcoming, numerous studies attempted to train the GAN with limited data, which can be divided into three categories. The prior-knowledge-based approaches \cite{ref12,ref13} introduce external data as an alternative. They first learn a semantically related generation and then adapt it to the generation of interest. Despite the effectiveness, additional images need to be collected same as mentioned above. Several data augmentation schemes tailored for GAN have been developed to alleviate the need for additional datasets. They use groups of data augmentation (e.g., rotation, crop) to transform the inputs of the discriminator \cite{ref15,ref16,ref33}. These methods avoid overfitting by applying regularization to the discriminator, generating high quality images even with limited data. Yet they have the risk of leakage and need to design complex transformations for different datasets. Interpolation-based approaches \cite{ref17,ref18}, which generate images by mixing features in an encoder's bottleneck, remain heavily reliant on large datasets for training. Furthermore, their architecture necessitates real images as input instead of sampling from a Gaussian distribution. This characteristic is not conducive to deployment in flexible downstream tasks, such as incremental learning or transferring the GAN, where control via a simple latent space is often required. Therefore, our design philosophy is twofold: to effectively regularize the discriminator to prevent overfitting in data-scarce settings, while deliberately maintaining methodological simplicity and avoiding the introduction of any potential artifacts or biases inherent in data augmentation strategies. In contrast, our objective is to achieve robust discriminator regularization through a minimalist mechanism, one that is intentionally free from complex data augmentation strategies\cite{ref27} to ensure that no external biases are introduced into the generation process.
\begin{figure*}[! t]
	\centering
	\includegraphics[width=0.9\textwidth]{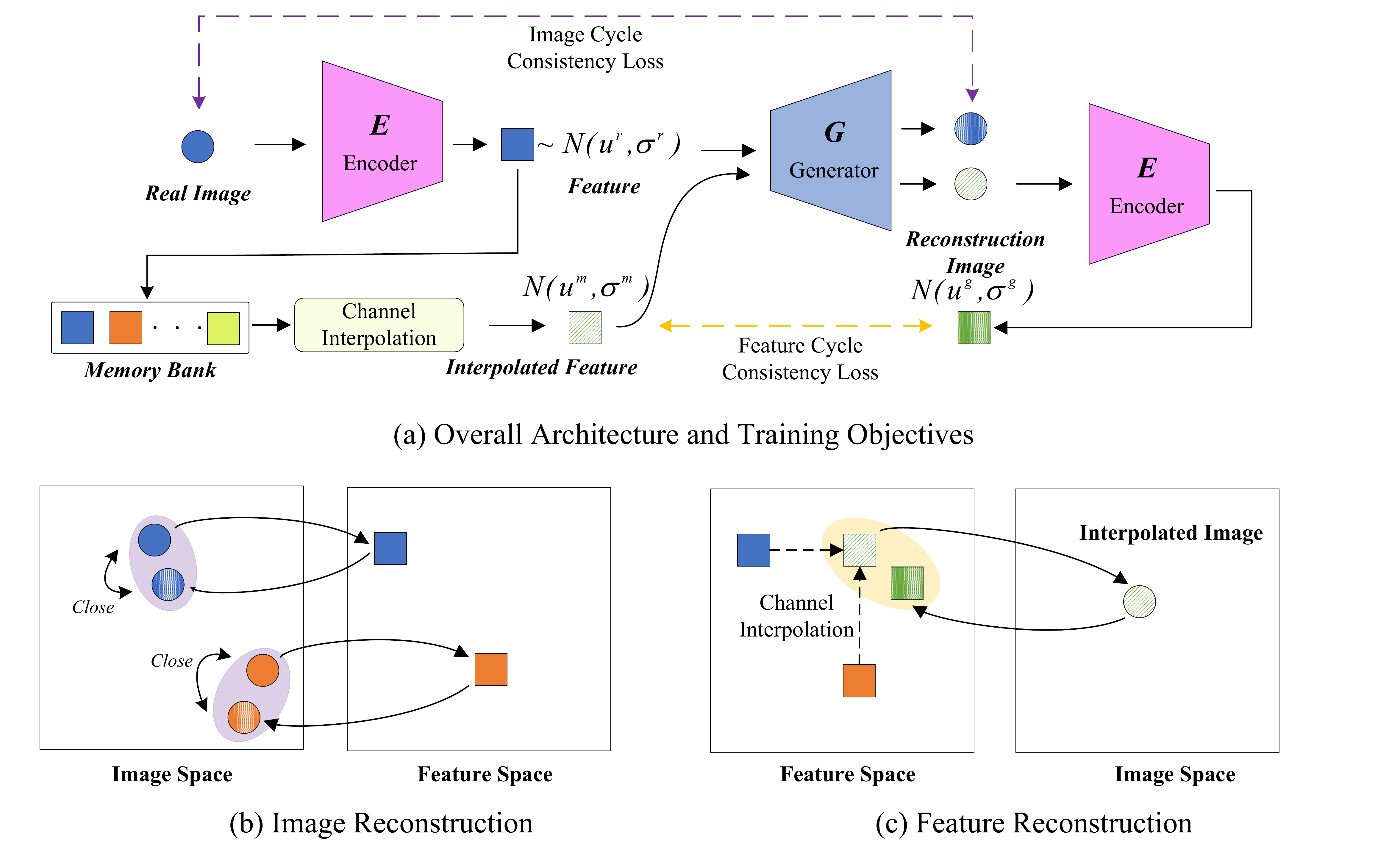}
	\caption{(a) The overall architecture, illustrating the data flow for both image reconstruction and feature interpolation. A real image is encoded into a feature, which can be either reconstructed directly or interpolated with other features from a memory bank to synthesize novel images. (b) Conceptual illustration of the Image reconstruction, which enforces that a reconstructed image should be close to its real-world counterpart. (c) Conceptual illustration of the Feature reconstruction, which regularizes the feature space by encouraging meaningful representations when interpolating between features.}
	\label{f1}
\end{figure*}

In this paper, we introduce a novel framework, consistency-regularized generative adversarial Network (Cr-GAN), which draws inspiration from Variational Autoencoders\cite{ref20} and the cycle-consistency mechanism of CycleGAN\cite{ref19}. We follow the idea of regularization of the discriminator, constraining the model through decoupling and reconstruction instead of augmentation. Specifically, a novel two-branch discriminator is presented. A branch is used for adversarial training with the generator, while the images are encoded by the other branch. The representation of real images will be interpolated randomly to obtain synthetic feature. Synthetic feature and real feature will be both input to the generator, which generates synthetic images and reconstructed images. The reconstructed images need to be close to the real images, and the synthetic image needs to be as realistic as possible. This is similar to the process from the source domain to the target domain of Cycle-GAN. The next step is to send the synthesized image to the discriminator to acquire reconstructed features. The reconstructed features should be closed to the synthetic features, and this is similar to the process from the target domain to the source domain of Cycle-GAN. In Fig. \ref{f1}, the overview of Cr-GAN is presented. In addition, we train the discriminator in VAE style \cite{ref20,ref39}. Therefore, in the generation stage, the input of generator is the samples from a gaussian distribution instead of images, which can be effectively generalized to downstream tasks. By imposing a consistency constraint in both the image and feature spaces, the model avoids common pitfalls of traditional GANs, such as mode collapse and overfitting, especially when trained on limited data.\par

We demonstrate the effectiveness of Cr-GAN on SAR datasets, showing that our method can effectively augment small labeled datasets with diverse and realistic synthetic images. The main contributions of this work can be summarized as follows:
\begin{enumerate}
	\item We propose Cr-GAN, an efficient generative framework tailored for few-shot SAR target recognition. Our core innovation lies in a dual-domain cycle consistency mechanism, which stabilizes training and enhances feature learning directly from limited data, obviating the need for external augmentation.
	
	\item A novel feature-space interpolation strategy for data scarcity. This approach synthesizes a rich spectrum of novel and diverse SAR samples by interpolating latent representations at the channel level, unlocking fine-grained, controllable generation that originates from simple Gaussian noise.
	
	\item We architect a two-branch discriminator that performs a dual-validation role. One branch assesses the quality of images from the baseline GAN pipeline, while the other is dedicated to enforcing consistency and realism for the novel samples generated through our feature interpolation mechanism.
	
	\item Extensive experiments on the MSTAR dataset validate the effectiveness of our method, which achieves $71.21\%$ accuracy under 8-shot settings, outperforming all GAN-based and diffusion-based methods, with significantly low computational cost.
\end{enumerate}

\section{Related Work}
This section reviews three core areas relevant to our work. We begin by surveying the evolution of SAR target recognition, from foundational Convolutional Neural Networks (CNNs) to modern Transformer-based architectures, highlighting a common bottleneck: the scarcity of labeled data. Subsequently, we examine dominant paradigms for training with limited data, including transfer learning, few-shot learning, and self-supervised methods. Finally, we delve into GAN, a key technology for data synthesis, with a specific focus on the challenges they face in data-scarce regimes and the limitations of existing solutions.
\subsection{Synthetic Aperture Radar target Recognition}
In recent years, deep learning has become the dominant methodology for SAR image classification. The application of deep CNNs, particularly the adaptation of seminal architectures like VGG-Net\cite{vgg} and ResNet\cite{resnet}, demonstrated a significant performance advantage over traditional methods by automatically learning hierarchical feature representations from SAR data \cite{sar1}. Building on this foundation, a multitude of advanced network designs have been proposed to further enhance classification accuracy. These include complex-valued CNNs (CV-CNNs)\cite{sar3} designed to process the raw phase information inherent in complex SAR data, and networks incorporating multi-scale feature fusion to capture target signatures at various resolutions \cite{sar4}. To improve feature discriminability amidst speckle noise and clutter, other works have integrated attention mechanisms, such as the Squeeze-and-Excitation (SE) module \cite{sar5} and the Convolutional Block Attention Module (CBAM) \cite{sar6}, into the network backbone.\par
Seeking to overcome the inherent limitations of local receptive fields in CNNs, recent research has increasingly explored Transformer-based architectures, which leverage self-attention to model long-range spatial dependencies. The Vision Transformer\cite{ViT-Base} (ViT) was among the first to be adapted for SAR classification, showing strong potential in capturing global contextual information. To improve computational efficiency and learn hierarchical representations, variants like the Swin Transformer\cite{Swin} have been successfully applied, achieving state-of-the-art results on public benchmarks. Furthermore, hybrid models that combine the local feature extraction capabilities of CNNs with the global modeling power of Transformers have emerged as a promising research direction, seeking to synergize the strengths of both architectural paradigms.\par
Despite the continuous advancement of these recognition models, their full potential is often constrained by the limited availability of labeled SAR data, a common challenge that can lead to model overfitting and poor generalization. Our work aims to address this critical data-scarcity issue. We introduce Cr-GAN, a novel generative model designed to supplement training sets by synthesizing high-fidelity and semantically diverse SAR images. The primary objective is to enrich the available data, thereby enhancing the performance and robustness of the aforementioned advanced classification models, particularly in data-scarce scenarios.

\subsection{Limited Data Training}
The challenge of training deep learning models, including those for SAR target recognition, is significantly exacerbated by the scarcity of labeled data, often leading to overfitting and poor generalization. To address this, several powerful paradigms have been developed. A primary strategy is transfer learning\cite{lml1,lml2}, which leverages knowledge from data-rich source domains. This is commonly realized by fine-tuning models pre-trained on large benchmarks, or more recently, through SSL methods\cite{ref1,ref2,ref3,ref4,ref6,ref8,lab1,lab2}, which learn robust representations from unlabeled data that adapt efficiently to downstream tasks. \par
A different paradigm, FSL\cite{lml4,TANG2,TANG3,TANG4,ref5,lml6,dba,hmn,dasp,uil}, directly tackles the problem of learning from a handful of examples. Metric-based FSL methods, such as Prototypical Networks\cite{lml3}, learn an embedding space where classification is performed by comparing new samples to class prototypes derived from the few available support examples. BIDFC\cite{WCL}, although successful in small sample settings, still requires extensive training with unlabeled data. While these approaches effectively utilize existing or external knowledge, a complementary strategy is to directly augment the scarce training data itself. Generative models, particularly Generative Adversarial Networks and Diffusion Models\cite{ref43,ref44,SiT,edm2,det4sar,madinet,jhdp}, offer a promising avenue for this by synthesizing novel, realistic data points to enrich the training set. We develop Cr-GAN to generate diverse samples through latent feature interpolation and dual consistency constraints, which serve as effective supervision for pretraining and improve downstream performance in few-shot SAR recognition.
\subsection{Generative Adversarial Networks}
The field of generative modeling has evolved rapidly, with GANs long serving as a cornerstone. Research in GANs has historically progressed along two main axes: the design of the loss function to stabilize training, evolving from the original JS-divergence \cite{ref14} to the Wasserstein distance \cite{ref22, ref23}, and the advancement of network architectures, such as StyleGAN \cite{ref26, ref27}, to enhance generation fidelity. More recently, Diffusion Models, including prominent examples like DDPM \cite{ref43} and the Transformer-based DiT \cite{ref44}, have emerged as the dominant paradigm, often setting new benchmarks in high-fidelity image generation. However, their substantial computational cost during the iterative sampling process remains a significant drawback. This has fueled renewed interest in improving the efficiency and quality of GANs, with recent efforts like R3GAN \cite{ref45} seeking to bridge this performance gap by revisiting GAN training stability.\par
\begin{figure*}[! t]
	\centering
	\includegraphics[width=\textwidth]{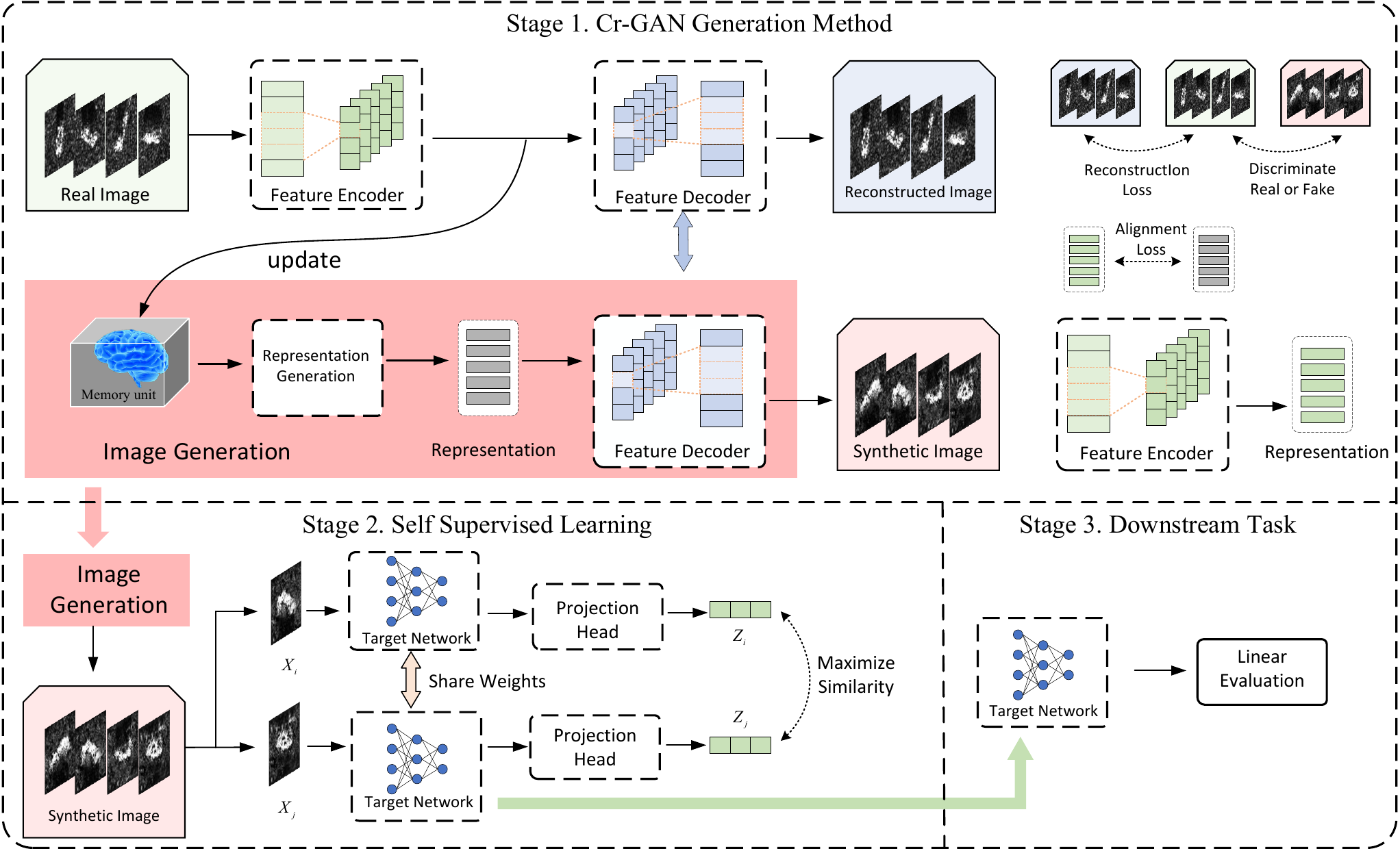}
	\caption{The proposed pipeline, built upon Cr-GAN and self-supervised learning, comprises three key stages: 
		(1) Cr-GAN Generation Method: We first train our Cr-GAN model on the limited real data. The objective is to learn the underlying data distribution and synthesize a large-scale, diverse dataset for the subsequent pre-training stage. 
		(2) Self-Supervised Learning: We leverage the synthetic dataset to pre-train a backbone network. By employing a contrastive learning objective, the model learns robust and generalizable feature representations without requiring any labels. 
		(3) Downstream Task: The pre-trained backbone is fine-tuned on the original, scarce set of labeled images, adapting the learned representations to the specific downstream classification task.
	}
	\label{pipeline}
\end{figure*}
A central challenge for generative models, particularly GANs, is their significant performance degradation in limited data settings, where the discriminator is prone to overfitting and memorizing the training set \cite{ref28, ref29}. Prevailing strategies to combat this either rely on transfer learning from large-scale source domains \cite{ref12, ref13, ref34} or apply strong regularization through extensive data augmentation \cite{ref15, ref31, ref33}. However, these solutions introduce their own dependencies. Transfer-based methods are contingent on the availability of a suitable source domain, while augmentation-heavy techniques risk leaking artifacts into the generated images. Furthermore, other methods that operate in the feature space often rely on pre-trained encoders from datasets like ImageNet \cite{ref35}, which may not generalize well to domains beyond natural images \cite{ref17, ref18}. In this paper, we design a new regularization task without data augmentation to avoid leakage. At the same time, the method does not require auxiliary knowledge, which can adapt to various generation tasks under limited data.

\section{Consistency Regularized Generative Adversarial Networks}
\begin{figure*}[! t]
	\centering
	\includegraphics[width=0.9\textwidth]{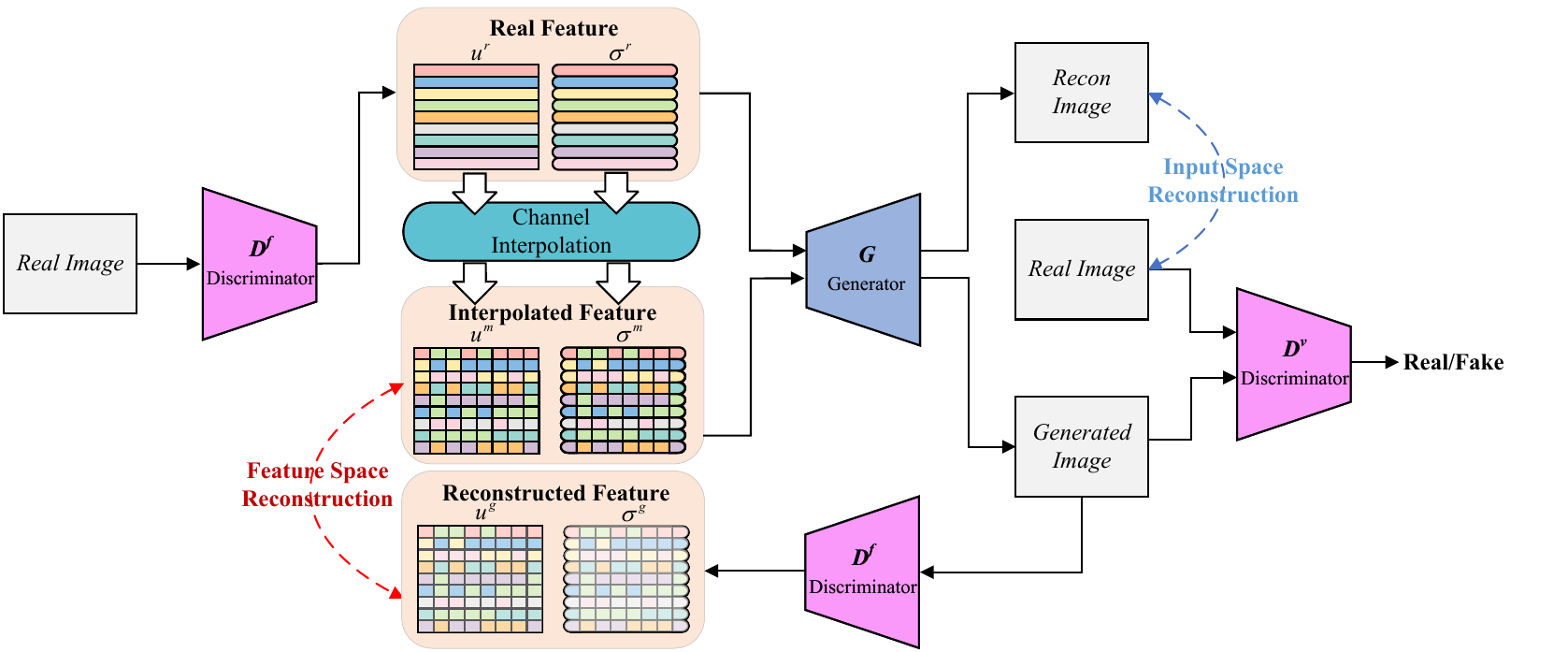}
	\caption{The proposed Cr-GAN framework. The discriminator's feature branch encodes real images into latent features. Channel-wise interpolation then synthesizes mixed features. The generator uses both original and mixed codes to produce reconstructed and novel synthetic images, respectively. Two losses provide supervision: the image reconstruction ensures fidelity, and the feature reconstruction aligns the generated features with the mixed features to ensure consistency. The superscripts $r, m, g$ denote real, mixed, and generated data.}
	\label{arc}
\end{figure*}

The architecture of our Cr-GAN generation method is illustrated in Fig. \ref{arc}. Given an image dataset $X$, we would like to learn a generator $G$ and discriminator $D$ via optimizing adversarial loss $\mathcal{L}_{GAN}$ that synthesizes diverse images under limited data. The discriminator has two branches, one branch used for adversarial training is denoted as $D^{v}$, another branch denoted as $D^{f}$ for representation, which outputs mean $\mu$ and $\sigma $ like VAE. The discriminator branch $D^{f}$ encode $x^{r}$ to obtain $\mu^{r}$ and $\sigma^{r}$ firstly where $x^{r}\in X$. We mix the output from real images to get $\mu^{m}$ and $\sigma^{m}$. The $z^{r}$ and $z^{m}$ is obtained by reparameterizing the mean and variance from the real images and mixture. The generator $G$ takes $z^{r}$ and $z^{m}$ as input to synthesize reconstructed images $\hat{x}$ and generated images $x^{g}$ respectively. We denote the data distribution as $x^{r}\sim P_{data}(x^{r})$ and generator's distribution as $x^{g}\sim P_{g}(x^{g})$. We minimize the image cycle consistency loss $\mathcal{L}_{IR}$ make the reconstructed image $\hat{x}$ close to the real image $x^{r}$. Meanwhile we send the generated data to the $D^{f}$ to obtain reconstructed mixture feature $\hat{\mu}$ and $\hat{\sigma}$. Finally, the feature cycle consistency loss $\mathcal{L}_{FR}$ is minimized. The key innovation of the proposed network lies in cycle consistency regularization between image and feature domain instead of data augmentation, which effectively avoids leakage and the overfitting of discriminator. Besides, in the generation stage, the generator still can accept samples from Gaussian distribution as input. After generating synthetic samples using our proposed method, we perform self-supervised pretraining on the generated data, followed by fine-tuning on the limited set of real labeled samples. The overview of our framework is illustrated as shown in Fig.\ref{pipeline}.

\setlength{\textfloatsep}{5pt}
\begin{figure}[!t]
	\centering
	\includegraphics[width=0.65\linewidth]{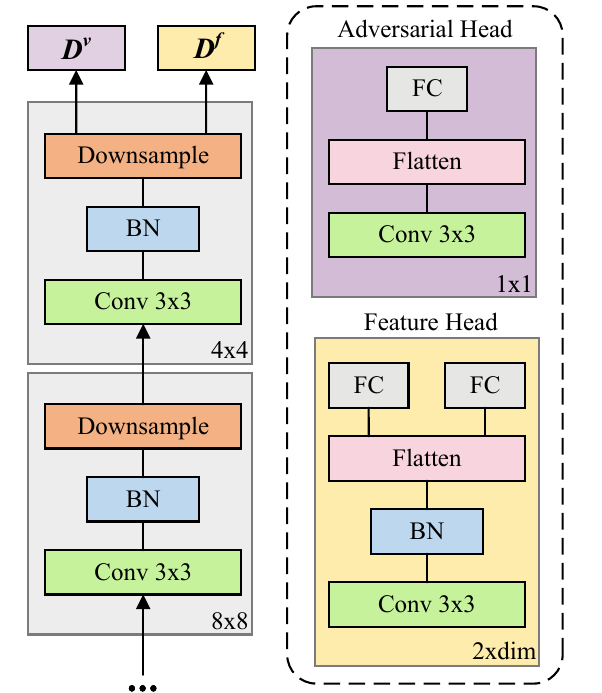}
	\caption{Two-branch discriminator architecture. A shared convolutional body extracts features that are fed into two distinct heads: an adversarial head for real/fake classification and a feature head that outputs the mean and variance.}	
	\label{architecture}
\end{figure}
\subsection{Preliminary: VAE-GAN}
Inspired by Larsen \cite{vaegan}, we apply VAE-GANs to the consistency on image domain. Unlike them, a two-branch discriminator $D^{f}$ is regularized by imposing a prior over latent distribution $p(z)$. Typically $z\sim N(0,1)$ is chosen. The latent representation from the real sample or generated samples can be written $z\sim q(z|x)=\mu +\varepsilon \sigma$, where $\mu$ and $\sigma$ are encoded by $D^{f}$, and $\varepsilon$ is an auxiliary noise variable $\varepsilon \sim N(0,1)$. Besides, we denote the input of generator to synthesize generated images as $z^{m}\sim P_{m}(z^{m})=\mu ^{m}+\varepsilon \sigma ^{m}$, where $\mu^{m}$ and $\sigma ^{m}$ is mixed from $\mu$ and $\sigma$, $\text{KL}$ is the Kullback-Leibler divergence, more details are illustrated in next section. The training objectives of VAE-GAN loss is the image reconstruction loss, adversarial loss and prior regularization term:
\begin{equation}
	\mathcal{L}^{G}_{IR}=\mathbb{E}_{x^{r}\sim P_{data}(x^{r}),z^{r}\sim q(z^{r}|x^{r})}[\left \| G(z^{r})-x^{r} \right \|_{1} ] \label{eq32},
\end{equation}
\begin{equation}
	\begin{aligned}
		\mathcal{L}^{D}_{GAN}= & -\mathbb{E}_{x^{r}\sim P_{data}(x^{r})}[D^{v}(x^{r})] \\ & + \mathbb{E}_{z^{m}\sim P_{m}(z^{m})}[D^{v}(G(z^{m}))] \label{eq33},
	\end{aligned}
\end{equation}
\begin{equation}
	\mathcal{L}^{G}_{GAN}=-\mathbb{E}_{z^{m}\sim P_{m}(z^{m})}[D^{v}(G(z^{m}))] \label{eq34},
\end{equation}
\begin{equation}
	\begin{aligned}
		\mathcal{L}_{prior}^{D}= &KL(q(z^{r}|x^{r})\left |  \right |p(z) ) \\ & +KL(q(z^{g}|x^{g})||p(z))  \label{eq35},
	\end{aligned}
\end{equation}
where $||\cdot||_{1}$ is the L1 norm, the generator $G$ and the discriminator branch $D^{f}$ try to achieve a good mutual conversion between the image domain and the feature domain. Since a large number of meaningful generated images through the translation of the feature domain to the image domain.
\subsection{Two-Branch Discriminator}
The discriminator plays two roles namely in representation and identification, the former depends on the non-linear capability of the model. However, the goal of identification task requires the discriminator and generator to approximate Nash equilibrium, meaning that the discriminator satisfies the Lipschitz continuity. However, non-linearity and smoothness has conflict. To solve this issue, we start from the discriminator architecture.

\textbf{Architecture}. We adopt the structure of two branches for task independence in structure to balance the independence and relevance of the two tasks. Meanwhile, using skip connections to make the feature reconstruction task regularize the main task. Take the DCGAN\cite{ref24} discriminator as an example, the structure of the discriminator is shown in Fig. \ref{architecture}.

\subsection{Consistency on the Discriminator}
The prior regularization term $\mathcal{L}_{prior}^D$ encourages the latent features to be independent of each other. Consequently, to synthesize novel and semantically meaningful representations, we mix the intermediate features extracted by our discriminator branch $D^f$. This mixing is performed via a process we term Channel Interpolation ($CI$), which operates at the feature level.

Specifically, for any two real feature, $x$ and $y$, $CI$ synthesizes a new feature by generating a simple random binary mask $K$, which dictates an element-wise selection from either input. The operation is written as:
\begin{equation}
	CI(x, y, K) = K \odot x + (1 - K) \odot y,
	\label{eq:ci_interpolation}
\end{equation}
where $x, y \in \mathbb{R}^N$ are the input feature vectors of dimension $N$, $K \in \{0, 1\}^N$ is the random binary mask, and $\odot$ denotes the Hadamard  product. We apply this operation to the mean and variance features extracted from two random source images, $x_1$ and $x_2$. The resulting interpolated features, $\mu^m$ and $\sigma^m$, are given by:
\begin{align}
	\mu^m &= CI(\mu_1^r,\mu_2^r,K) \label{eq:mu_interp}, \\
	\sigma^m &= CI(\sigma_1^r,\sigma_2^r,K) \label{eq:sigma_interp},
\end{align}
where $(\mu_1^r, \sigma_1^r) = D^f(x_1)$ and $(\mu_2^r, \sigma_2^r) = D^f(x_2)$.\par
The validity of Channel Interpolation is grounded in the inherent properties of VAE latent spaces and the Manifold Assumption. Specifically, our discriminator $D^f$ is regularized via a standard VAE objective with a diagonal Gaussian prior (Eq.~\ref{eq35}). $\beta$-VAE~\cite{betavae} shows that this independence constraint encourages latent channels to encode distinct semantic variations of the data. Consequently, swapping channels operates as a composition of semantic attributes (e.g., combining the scattering topology of one target with the background texture of another), rather than arbitrary noise mixing. Regarding physical validity, recently, ~\cite{JIT} pointed out the phenomenon that real data resides on a low-dimensional manifold. Since our generator is trained to reconstruct real images, it effectively learns the mapping from the latent space to this valid manifold. Thus, even if a mixed latent code $z^m$ theoretically lies slightly off-manifold, the generator acts as a projector, mapping it back to a physically valid SAR image that adheres to the underlying geometric and scattering constraints.

\begin{figure}[! t]
	\centering
	\includegraphics[width=0.9\columnwidth]{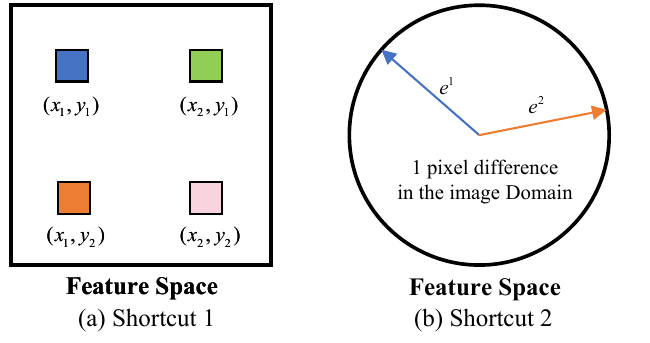}
	\caption{Illustration of shortcut learning. 
		(a) When synthesizing a new feature by interpolating two existing ones, a naive model might learn a trivial mapping instead of generating a semantically novel representation.
		(b) A single-pixel variation in the image space can correspond to a substantially amplified discrepancy in the feature space, indicating a non-robust representation.
	}
	\label{shortcut}
\end{figure}

For the interpolated inputs to be useful for training, we first reparamtrize the mixture feature $\mu^{m}$ and $\sigma^{m}$ to obtain $z^{m}$, and apply to the generator to synthesize the generated images $x^{g}$, i.e., $x^{g}=G(z^{m})$. To achieve the goal of generation under limited data, the decoupling and reconstruction regularization in the feature space is introduced. We enforced the features of generated images encoded from the discriminator branch $D^{f}$ to close to the mixture feature. We denote $\mu^{g}\sim q(\mu^{g}|x^{g})$, $\sigma^{g}\sim q(\sigma^{g}|x^{g})$, $\mu^{m}\sim P_{m}(\mu^{m})$ and $\sigma^{m}\sim P_{m}(\sigma^{m})$. The target is minimized by the feature cycle consistency loss $\mathcal{L}^{D}_{FR}$, which is written as:
\begin{equation}
	\begin{aligned}
		\mathcal{L}^{D}_{FR}= & \mathbb{E}_{\mu^{g}\sim q(\mu ^{g}|x^{g}),\mu ^{m}\sim P_{m}(\mu ^{m})}d(\mu ^{g}, \mu ^{m}) \\ & + \mathbb{E}_{\sigma^{g}\sim q(\sigma ^{g}|x^{g}),\sigma ^{m}\sim P_{m}(\sigma ^{m})}d(\sigma ^{g}, \sigma ^{m}) \label{eq39},
	\end{aligned}
\end{equation}
where $d$ is the distance, which can be Euclidean distance, cosine similarity, etc. However, if Eq. \ref{eq39} is optimized for the regularization of discriminator, shortcut learning will be introduced. As shown in Fig. \ref{shortcut}.\par
\textbf{Shortcut 1.} Channel interpolation is essentially combining the representation of different samples to construct new features. The simplest way for discriminator to decouple and reconstruct the generated samples in the feature space is to make the mixed feature equivalent to the representation of the third sample. Therefore, we expect the goal of $\mathcal{L}^{D}_{FR}$ optimization have two-fold: 1) reconstruction and 2) uniform in the feature space, we write the Eq. \ref{eq39} as:
\begin{equation}
	\small
	\begin{aligned}
		\mathcal{L}^{D}_{FR} &= \mathbb{E}_{\substack{\mu^{g} \sim q(\mu^{g} | x^{g}) \\ \mu^{m} \sim P_{m}(\mu^{m})}} \big[ \mathcal{L}^{D}_{AU}(\mu^{g}, \mu^{m}, \{\mu^r_j\}_{j=1}^N) \big] \\
		&\quad + \mathbb{E}_{\substack{\sigma^{g} \sim q(\sigma^{g} | x^{g}) \\ \sigma^{m} \sim P_{m}(\sigma^{m})}} \big[ \mathcal{L}^{D}_{AU}(\sigma^{g}, \sigma^{m}, \{\sigma^r_j\}_{j=1}^N) \big] \label{eq40},
	\end{aligned}
\end{equation}
\begin{small} 
	\begin{equation}
		\label{eq:L_AU_single}
		\resizebox{0.9\hsize}{!}{$
			\mathcal{L}_{AU}(q, k^+, \mathcal{K}^-) = - \log \frac{e^{\text{sim}(q, k^+)/\tau}}{e^{\text{sim}(q, k^+)/\tau} + \sum_{k^- \in \mathcal{K}^-} e^{\text{sim}(q, k^-)/\tau}}
			$}
	\end{equation}
\end{small}

Where $sim(u, v)$ represents cosine similarity between image representations \( u \) and \( v \), $\tau$ is the temperature parameter and $\mathcal{L}^{D}_{AU}$ is the alignment uniform loss inspired by the InfoNCE\cite{ref38} loss of contrastive learning. $\{\mu^r_j\}_{j=1}^N$ and $\{\sigma^r_j\}_{j=1}^N$ denote the sets of mean and variance representations encoded by the discriminator branch $D^f$ from the current batch of real images. When Eq. \ref{eq40} is optimized by the discriminator, the representation in the feature space tends to uniform, avoiding trivial solution.

\textbf{Shortcut 2.} Eq. \ref{eq40} aims to repel the similar representation and make the mixture feature and generated features closer, yet the objective function makes the correlation of samples is not equivalent in the feature domain and the image domain. It brings second shortcut as illustrated in the Fig. \ref{shortcut} (b). To counteract this degeneracy and ensure that our Channel Interpolation strategy effectively produces diverse samples, we adopt the mode seeking loss proposed by \cite{ref41}. The equivalence about the correlation of samples in the feature and the image domain is achieved via increasing the diversity in the image domain, which can be formulated as:
\begin{equation}
	\mathcal{L}^{G}_{MS}=-\mathbb{E}_{z_{1},z_{2}\sim q(z^{r}|x^{r})\cup P_{m}(z^{m})}\frac{\left \| G(z_{1})-G(z_{2}) \right \|_{1} }{\left \| z_{1}-z_{2} \right \|_{1} } \label{eq42}.
\end{equation}

Maximizing this distance ratio functions as a regularizer that imposes a lower bound on the generator's Lipschitz constant. This directly counteracts the root cause of ``Shortcut 2'', the degeneracy of the mapping $G(z)$ where its Lipschitz constant approaches zero, leading to mode collapse. Consequently, the generator is compelled to learn an expansive mapping that is highly sensitive to latent variations, thus preserving the crucial correlation between the latent and image domains.
\par
\textbf{Full Objective.} The algorithm can be divided into generator and discriminator training written as:
\begin{align}
	\mathcal{L}^{D} &= \lambda_{GAN}\mathcal{L}^{D}_{GAN}+ \lambda_{feat}\mathcal{L}^{D}_{FR}+\lambda_{pr}\mathcal{L}^{D}_{prior} \label{eq43}, \\
	\mathcal{L}^{G} &= \lambda_{GAN}\mathcal{L}^{G}_{GAN}+\lambda_{IR}\mathcal{L}^{G}_{IR}+\lambda_{ms}\mathcal{L}^{G}_{MS} \label{eq44}.
\end{align}
\par
The synergy between our channel interpolation and dual-domain cycle consistency serves as a powerful internal regularizer that prevents the discriminator from overfitting in the few-shot regime. The channel interpolation acts as a form of implicit data augmentation in the feature space, creating a continuous distribution of novel yet plausible latent samples. This forces the discriminator to learn a generalized decision boundary rather than simply memorizing the small set of real images. The cycle consistency, particularly the feature-cycle consistency ($z^m \rightarrow x^g \rightarrow z^g$), provides the crucial structural constraint for this process. It imposes a self-supervised task on the discriminator's encoder branch, forcing it to learn a semantically smooth and consistent mapping of the generative space. In essence, interpolation creates a diverse set of challenging examples, while consistency provides the underlying rules for learning from them, jointly regularizing the discriminator to learn robust and generalizable features.

\subsection{Self-Supervised Learning Framework}

In this study, we adopt the SimCLR\cite{ref1} framework for SSL to pretrain a model using a large number of unlabeled images. The core idea of SimCLR is to learn image representations by bringing similar images (positive pairs) closer in the feature space while pushing dissimilar images (negative pairs) farther apart. We use the InfoNCE\cite{ref38} loss, which encourages the model to distinguish between positive and negative samples, as detailed in the following equation:

\begin{equation}
	l_{i,j} = -\log \frac{\exp(\text{sim}(z_i, z_j)/\tau)}{\sum_{k=1}^{2N} 1_{[k \neq i]} \exp(\text{sim}(z_i, z_k)/\tau)},
\end{equation}
where \( \tau \) is the temperature parameter that controls the concentration of the similarity, \(1_{[k \neq i]} \) ensures that the negative samples are selected from all other instances except the positive pair.

For pretraining, we construct a dataset generated by Cr-GAN comprising 5000 synthetic images. This strategy significantly enhances the model's generalization capabilities, particularly in scenarios where labeled data is scarce. To further augment the robustness of the learned features, we apply a series of geometric and photometric transformations to the input images, including random cropping, color jittering, and horizontal flipping. These transformations ensure that the network develops invariant representations that are resilient to common variations in imaging conditions.

Following the pretraining phase, the model is fine-tuned on a downstream task using a comparatively smaller labeled dataset. This transfer learning approach capitalizes on the rich feature representations acquired during the self-supervised pretraining stage, thereby improving performance on tasks characterized by limited labeled data, such as image classification or object detection in RS applications.

\section{Experiments}
\subsection{Dataset}
\textbf{MSTAR}. The Moving and Stationary Target Acquisition and Recognition (MSTAR)\cite{mstar} dataset contains 5,950 annotated 64×64 pixel samples of military targets spanning ten distinct operational categories, acquired through 0.3-meter resolution spotlight SAR. Training samples were collected at 17° depression angle while test samples were captured at 15° depression angle.\par
\begin{figure}[! t]
	\centering
	\includegraphics[width=\columnwidth]{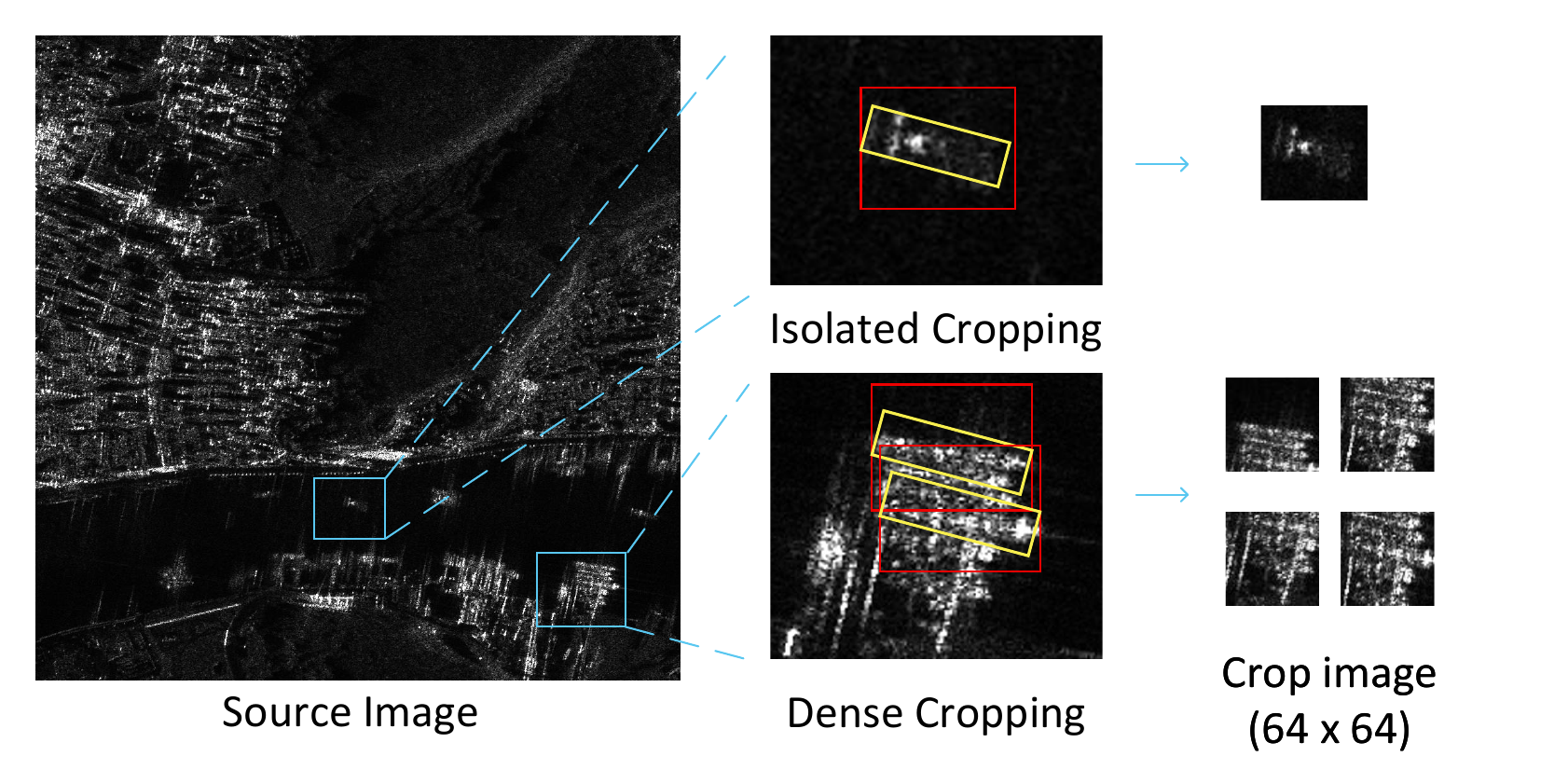}
	\caption{The data processing pipeline for adapting the SRSDD detection dataset. The minimum enclosing square patch is cropped for each target. While this is straightforward for isolated targets, it introduces noise for dense targets, creating a more challenging classification task. In the magnified views, the yellow rotated rectangles represent the ground-truth labels from the detection task, while the red squares indicate the final cropped image patches.
	}
	\label{srsdd}
\end{figure}

\textbf{SRSDD}. To further assess the generalization of Cr-GAN, we conducted additional experiments on the public SRSDD-v1.0 SAR ship detection dataset \cite{SRSDD}. SRSDD is a challenging, high-resolution benchmark with six distinct ship categories. As illustrated in the Fig \ref{srsdd}. We adapted this detection dataset for our few-shot classification task by cropping the minimum enclosing square patch for each annotated target. A key characteristic of this process is that the close proximity of ships in port scenes often results in cropped patches containing multiple or partial objects. This inherently introduces label noise, providing a more challenging and pragmatic test of a model's robustness.

\subsection{Implementation Details}
All experiments can be divided into three parts: image generation,model pretraining and downstream Fine-tuning. In former process, we use DCGAN\cite{ref24} as the network architectures to verify the generalization of proposed method, it is noted that all hyperparameters adhere to original setting. In our experiment settings, temperature coefficient is set to $0.2$, and we set the loss weights $\lambda_{\mathrm{GAN}}$, $\lambda_{\mathrm{pr}}$, and $\lambda_{\mathrm{IR}}$ to 1.0. The weights for the $\lambda_{\mathrm{feat}}$, and the $\lambda_{\mathrm{ms}}$ are set to 0.1, a value that is empirically validated as effective in our subsequent ablation study. The dimension of feature head in discriminator is the same as generator's. Momentum factor of memory bank is set to $0.999$ for weight updating. It is worth nothing that, We utilized only basic data augmentations (random rotation and flipping) without any background segmentation or clutter suppression preprocessing. For the self-supervised pre-training phase, we employed the SimCLR\cite{ref1} framework on our dataset of 5,000 synthesized images. The model was trained for 100 epochs. Key hyperparameters were set following the methodology of the original study\cite{ref1}.   In the finetuning period of downstream tasks, we choose Adam\cite{ref42} optimizer, total $100$ epochs, initial learning rate $0.003$, and operate learning-rate decay on 30th and 80th epochs, where decay ratio is $0.1$. For the downstream classifier fine-tuning stage, due to the scarcity of labeled samples, we utilize standard random horizontal flipping and rotating as a basic data augmentation technique to prevent the classifier from overfitting. Note that this augmentation is applied only during classifier training. The fine-tuning process was strictly confined to the original training samples for the generative network, deliberately excluding any synthetic data. It is noted that all experiments are carried on RTX 4080 GPU.

\subsection{Main Results}
\begin{table*}[htbp]
	\centering
	\caption{Few-Shot Classification Performance Comparison. Models are grouped by type. The following metrics are reported: Precision (\%), Accuracy (\%), Recall (\%), and F1-score (\%). The training iterations for generative models are also listed. The best result in each metric column is marked in \textbf{bold}, and the second-best result is \underline{underlined}. The results are averaged over 3 random seeds.}
	\label{tab:main_results}
	\resizebox{\textwidth}{!}{
		\begin{tabular}{l|l|cccc|cccc|cccc}
			\toprule
			{\textbf{Train Iter.}} & 
			{\textbf{Method}} & \multicolumn{4}{c|}{\textbf{2-shots}} & \multicolumn{4}{c|}{\textbf{4-shots}} & \multicolumn{4}{c}{\textbf{8-shots}} \\
			\cmidrule(lr){3-6} \cmidrule(lr){7-10} \cmidrule(lr){11-14}
			& & Precision & Accuracy & Recall & F1 & Precision & Accuracy & Recall & F1 & Precision & Accuracy & Recall & F1 \\
			\midrule
			\multicolumn{14}{l}{\textit{(Standard Fine-tuning model)}} \\
			\midrule
			- & Resnet18\cite{resnet} & 37.82 & 33.22 & 34.70 & 33.35 & 42.87 & 39.99 & 41.41 & 39.29 & 59.97 & 58.26 & 59.77 & 57.33 \\
			- & Resnet50\cite{resnet} & 27.14 & 25.98 & 22.55 & 19.00 & 29.03 & 36.34 & 32.04 & 27.22 & 60.34 & 60.72 & 59.80 & 58.91 \\
			- & resnext50\cite{resnext50} & 22.77 & 29.32 & 27.30 & 21.32 & 39.97 & 35.72 & 35.33 & 31.85 & 56.79 & 54.45 & 52.87 & 52.05 \\
			- & Resnext101\cite{resnext50} & 37.34 & 30.41 & 25.19 & 21.22 & 44.74 & 43.62 & 43.12 & 38.97 & 56.76 & 51.36 & 51.19 & 49.15 \\
			- & Efficientnet\cite{Efficientnet} & 29.19 & 25.60 & 27.46 & 25.55 & 48.85 & 38.46 & 39.02 & 36.25 & 51.15 & 50.39 & 49.26 & 48.04 \\
			- & Densenet121\cite{Densenet121} & 34.34 & 28.35 & 27.01 & 22.63 & 46.84 & 45.43 & 40.88 & 37.78 & 59.33 & 60.57 & 59.05 & 57.16 \\
			- & Convnext\cite{Convnext} & 29.88 & 26.54 & 27.25 & 22.82 & 22.88 & 29.82 & 20.19 & 13.97 & 34.21 & 38.65 & 33.28 & 29.77 \\
			- & Swin\cite{Swin} & 39.68 & 28.69 & 25.76 & 20.77 & 23.31 & 32.13 & 21.83 & 16.67 & 30.03 & 33.72 & 34.87 & 27.15 \\
			- & Swin\_v2\cite{Swin_v2} & 24.49 & 28.97 & 24.70 & 21.23 & 1.83 & 18.33 & 10.00 & 3.10 & 49.18 & 46.71 & 44.87 & 45.59 \\
			- & Regnet\cite{Regnet} & 34.68 & 36.75 & 31.96 & 29.78 & 32.87 & 36.99 & 31.43 & 25.46 & 51.78 & 52.67 & 49.90 & 48.45 \\
			- & ViT-Base(finetune-head)\cite{ViT-Base} & 15.89 & 16.23 & 19.11 & 13.05 & 22.09 & 20.89 & 20.96 & 20.89 & 22.55 & 23.17 & 26.00 & 21.56 \\
			- & ViT-Base\cite{ViT-Base} & 22.09 & 25.01 & 21.33 & 18.24 & 35.85 & 29.66 & 29.59 & 28.05 & 51.27 & 47.30 & 51.89 & 48.42 \\
			\midrule
			\multicolumn{14}{l}{\textit{(Diffusion-based Generative Methods)}} \\
			\midrule
			15k & DDPM\cite{ref43} & 44.60 & 40.68 & 39.79 & 38.69 & 56.15 & 52.45 & 52.30 & 52.25 & 65.20 & 70.55 & 65.10 & 65.15 \\
			15k & DiT(B/2)\cite{ref44} & 46.41 & 34.97 & 37.44 & 37.44 & 48.39 & 47.43 & 46.21 & 44.88 & 67.75 & 68.31 & 67.17 & 66.81 \\
			15k & DiT(S/2)\cite{ref44} & Fail & Fail & Fail & Fail & Fail & Fail & Fail & Fail & Fail & Fail & Fail & Fail\\
			15k & SiT(B/2)\cite{SiT}   & 52.19 & 45.46 & \underline{44.48} & \underline{43.97} & \textbf{60.65} & \underline{60.01} & \underline{55.64} & \underline{56.00} & \underline{68.15} & \underline{71.10} & \underline{67.50} & \underline{67.82} \\
			15k & SiT(S/2)\cite{SiT} & 36.85&33.41& 32.48 & 30.71 &40.36 &43.15&39.37&38.47 & 51.52 &50.98&48.55&49.11\\
			15k & EDM2\cite{edm2} & 50.41 & \underline{46.52} & 43.27 & 42.91 & 50.41 & 46.52 & 43.27 & 42.90 & 65.91 & 70.40 & 65.32 & 64.86\\
			15k & EDM2(With Pretrained)\cite{edm2} &\textbf{52.98}&43.83&46.28&43.87&57.45 &52.70&51.90&51.81&66.86&68.90&65.90&65.83\\
			\midrule
			\multicolumn{14}{l}{\textit{(GAN-based Generative Methods)}} \\
			\midrule
			15k & DCGAN\cite{ref24} & 43.58 & 36.49 & 35.23 & 35.10 & 52.80 & 51.63 & 49.76 & 48.73 & 64.92 & 68.50 & 64.58 & 63.68 \\
			15k & R3GAN\cite{ref45} & \underline{51.66} & 40.65 & 41.01 & 40.51 & \underline{56.50} & 57.04 & 51.05 & 52.27 & 62.47 & 62.47 & 62.31 & 61.79 \\
			15k & StyleGAN2\cite{ref27} & 45.82 & 43.43 & 42.33 & 39.74 & 43.45 & 47.17 & 42.58 & 41.29 & 62.98 & 65.16 & 61.65 & 61.77\\
			15k & StyleGAN2 + DiffAugment\cite{ref31}& 47.31 & 44.11 & 42.98 & 42.74 & 51.62 & 51.14 & 50.64 & 49.30 & 60.91 & 61.75 & 59.68 & 59.49\\
			15k & StyleGAN2 + DA\cite{ref31} + $R_{LC}$\cite{ref15} & 46.04 & 31.94 & 33.27 & 34.69 & 40.48 & 42.93 & 41.12 & 38.59 & 66.22 & 67.53 & 66.22 & 65.87 \\
			15k & StyleGAN2 + ADA\cite{ref33} & 41.98 & 38.46 & 39.41 & 35.79 & 53.04 & 56.85 & 53.35 & 52.63 & 68.13 & 69.06 & 67.08 & 67.41 \\
			\rowcolor{light-gray}\textbf{15k} & Cr-GAN(ours) & 48.10 & \textbf{47.24}{\tiny(+0.72)} & \textbf{46.42}{\tiny(+1.94)} & \textbf{47.20}{\tiny(+3.23)} & 58.40 & \textbf{60.06}{\tiny(+0.05)} & \textbf{58.25}{\tiny(+2.61)} & \textbf{57.37}{\tiny(+1.37)} & \textbf{72.50}{\tiny(+4.35)} & \textbf{71.21}{\tiny(+0.11)} & \textbf{70.01}{\tiny(+2.51)} & \textbf{71.23}{\tiny(+3.41)} \\
			\bottomrule
		\end{tabular}
	}
\end{table*}
\vspace{0pt}
\begin{figure}[! t]
	\centering
	\includegraphics[width=\columnwidth]{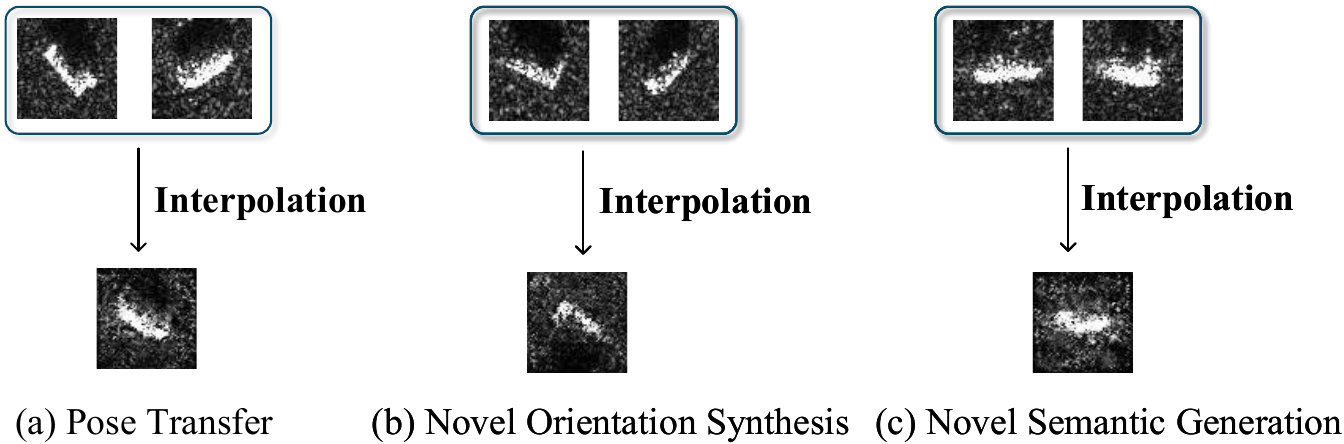}
	\caption{Visual demonstration of the semantic diversity generated by our Channel Interpolation strategy. 
		(a) Pose Transfer: Combining the orientation from the left source image with the content from the right source image. 
		(b) Novel Orientation Synthesis: By interpolating feature, the generator synthesizes a image at a completely new angle not present in the source inputs.
		(c) Novel Semantic Generation: It shows model generates unique structural features that do not exist in the training set, effectively enriching the diversity of the few-shot dataset.}
	\label{fig:interpolation_semantics}
	\label{interpolation}
\end{figure}
To rigorously evaluate our method, we conducted experiments under challenging few-shot conditions, using only 2, 4, and 8 labeled samples per class. 
For MSTAR\cite{mstar} results in Table~\ref{tab:main_results}, our approach consistently establishes a new state-of-the-art across all metrics. In the 8-shot setting, it achieves an impressive 71.21\% accuracy and a 71.23\% F1-score. This result not only substantially outperforms classic generative baselines like DCGAN\cite{ref24} but also surpasses the strongest modern competitors from both the GAN and Diffusion families, such as R3GAN\cite{ref45} and the recently proposed SiT\cite{SiT} and EDM2\cite{edm2} models. This significant performance vantage is not incidental; it originates from our method's unique generative process. This significant performance vantage is not incidental; it originates from our method's unique generative process. Instead of simply replicating existing data patterns, our model first extracts robust representations from the limited real samples. It then innovatively synthesizes new and varied features within this learned latent space creating representations that differ from the source samples before decoding them into a set of high-fidelity, diverse images. This generative capability is visually corroborated in Fig.~\ref{interpolation}. 

This ability to generate a dataset with superior diversity is the ideal prerequisite for the subsequent SSL stage. The rich visual variance in the synthetic images provides a perfect foundation for SimCLR\cite{ref1}, whose contrastive objective thrives on diverse data to learn a generalizable and semantically rich feature space. This robust pre-training provides an outstanding initialization for the downstream classifier, enabling remarkable data efficiency. This is clearly demonstrated by the model's performance trajectory: its accuracy skyrockets from 47.24\% to 71.21\% when moving from 2 to 8 shots per class, showcasing an unparalleled ability to capitalize on every available sample. In essence, our framework excels by first creating superior synthetic data through latent-space feature synthesis and then distilling its rich diversity into a powerful model representation via contrastive learning.

On the challenging SRSDD dataset~\cite{SRSDD}, as detailed in Table~\ref{tab:srsdd}, the superiority and robustness of Cr-GAN are particularly evident. We adhered to the standard evaluation protocol used for Table~\ref{tab:main_results} by reporting the average performance across three independent runs with different random seeds. Compared to other GAN-based methods, Cr-GAN establishes a significant performance margin across all few-shot settings. However, its data efficiency becomes most apparent when contrasted with modern Diffusion models.
\begin{figure}[! t]
	\centering
	\includegraphics[width=\columnwidth]{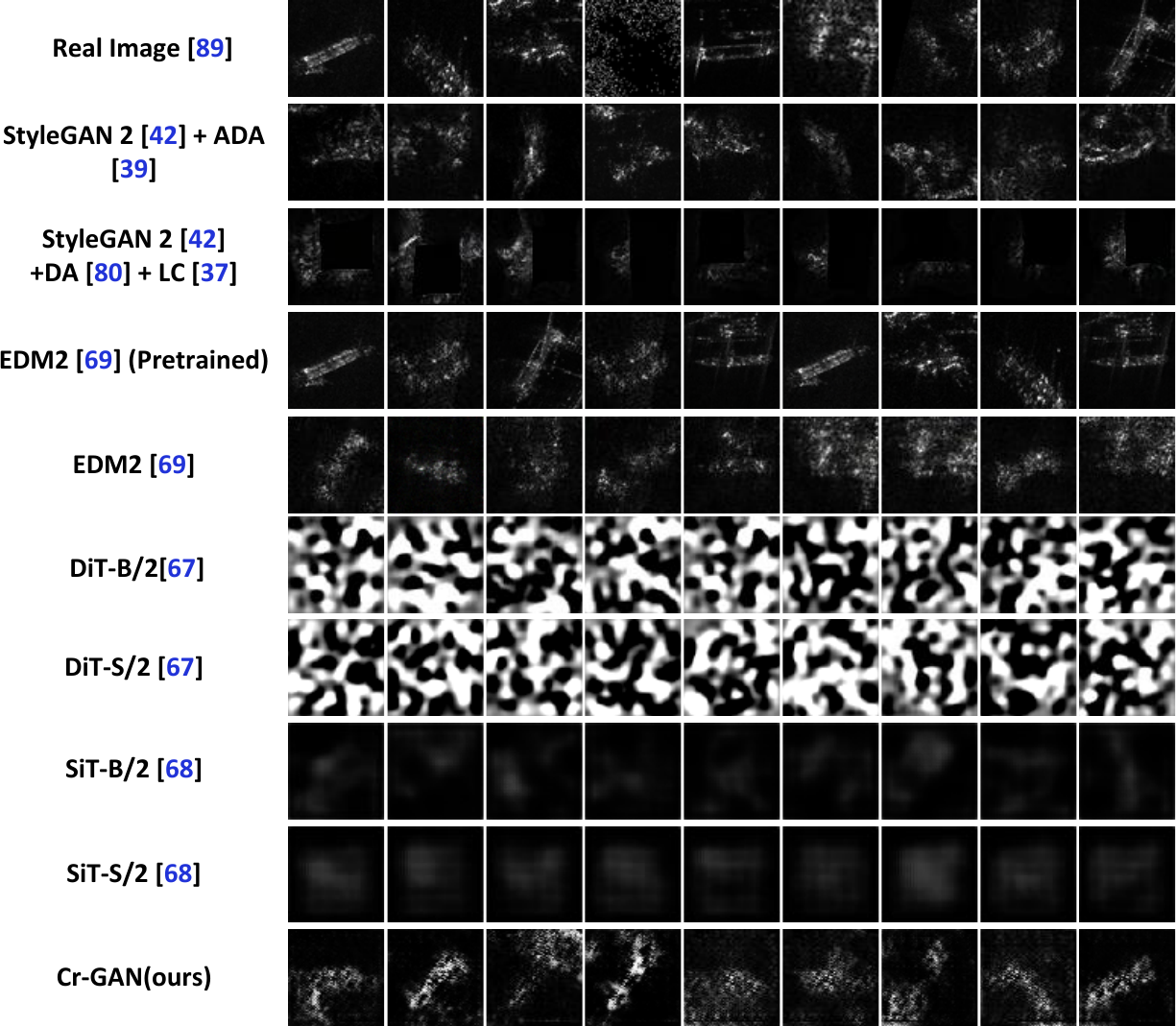} 
	\caption{Qualitative comparison of synthesized samples on the SRSDD dataset\cite{SRSDD} under the 8-shot setting. While our Cr-GAN  generates diverse samples with distinct structural variations, EDM2\cite{edm2} suffers mode collapse, producing images that are virtually replicas of the training data. In contrast, Transformer-based diffusion models (DiT\cite{ref44} and SiT\cite{SiT}) fail to learn meaningful representations, outputting noise or indistinct artifacts.}
	\label{srsdd_gen}
\end{figure}

A qualitative inspection, as presented in Fig.~\ref{srsdd_gen}, reveals distinct failure mechanisms among these baselines. In the extreme few-shot regimes, prominent Transformer-based architectures like DiT~\cite{ref44} and SiT~\cite{SiT} fail to converge entirely. Despite scaling down to 32M parameters, they consistently produce unusable noise or artifacts. The reasons are twofold, first, DiT lacks the inductive biases inherent in CNNs, making it struggle to capture structural patterns from scratch with limited samples. Second, SiT relies on learning a deterministic flow to map noise onto the data manifold. In data-scarce scenarios, the target manifold becomes sparse and ill-defined, preventing the model from identifying stable flow directions or trajectories, which results in chaotic, non-convergent outputs.

Regarding the advanced EDM2~\cite{edm2}, with optimized preprocessing, it successfully converged to generate realistic SAR images. However, a critical limitation emerged: despite visual fidelity, EDM2 suffers from severe mode collapse. Since diffusion models approximate the data probability density, with extremely few samples, the density collapses into discrete spikes. The model thus replicates the training samples with negligible variation, failing to provide the feature diversity necessary for effective pretraining.

This outcome underscores Cr-GAN's exceptional trade-off between accuracy and efficiency. It establishes Cr-GAN as a practical solution that circumvents not only the training instability of diffusion models in low-data regimes but also their inherently slow, iterative sampling process for image generation.
\begin{figure}[! t]
	\centering
	\includegraphics[width=\columnwidth]{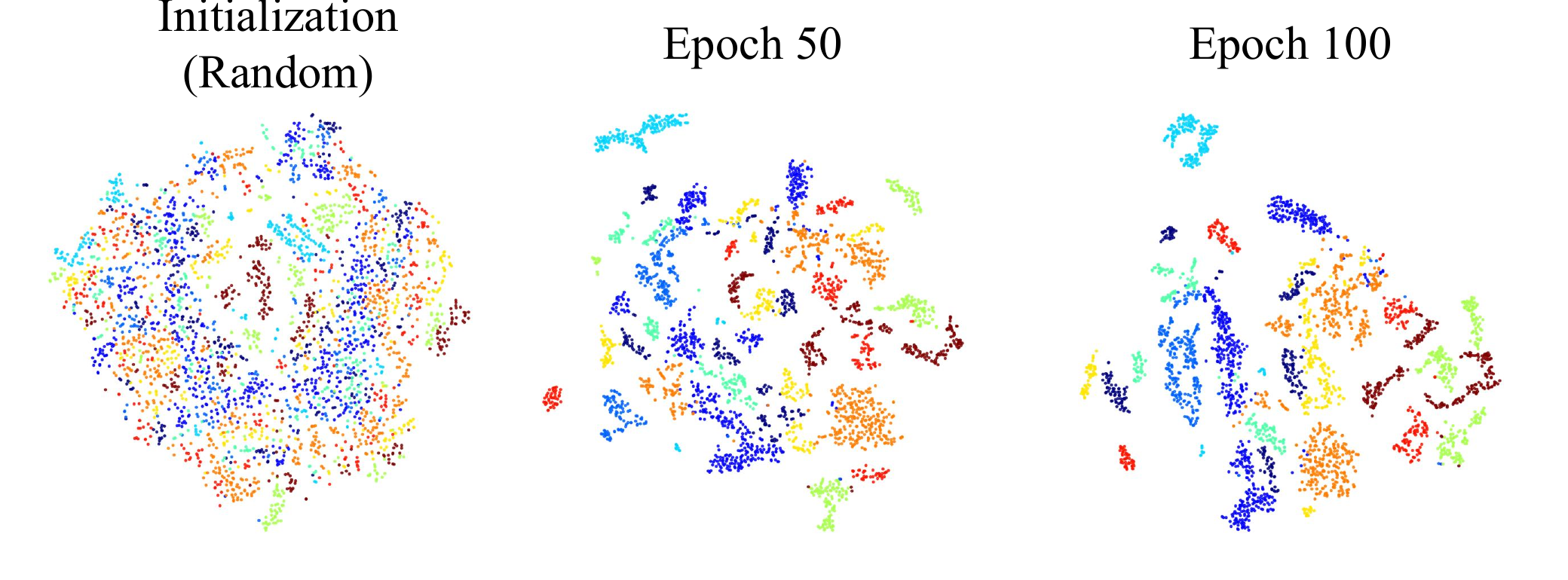} 
	\caption{Visualization of the dynamic evolution of the feature space learned by Cr-GAN during pre-training.
		Left: The process begins with a chaotic distribution from random initialization. 
		Middle: At Epoch 50, distinct class structures begin to emerge as the model learns from the synthetic data. 
		Right: By Epoch 100, the features form compact, well-separated clusters.}
	\label{fig:tsne_dynamics}
\end{figure}

\begin{figure}[!t]
	\centering
	\includegraphics[width=\columnwidth]{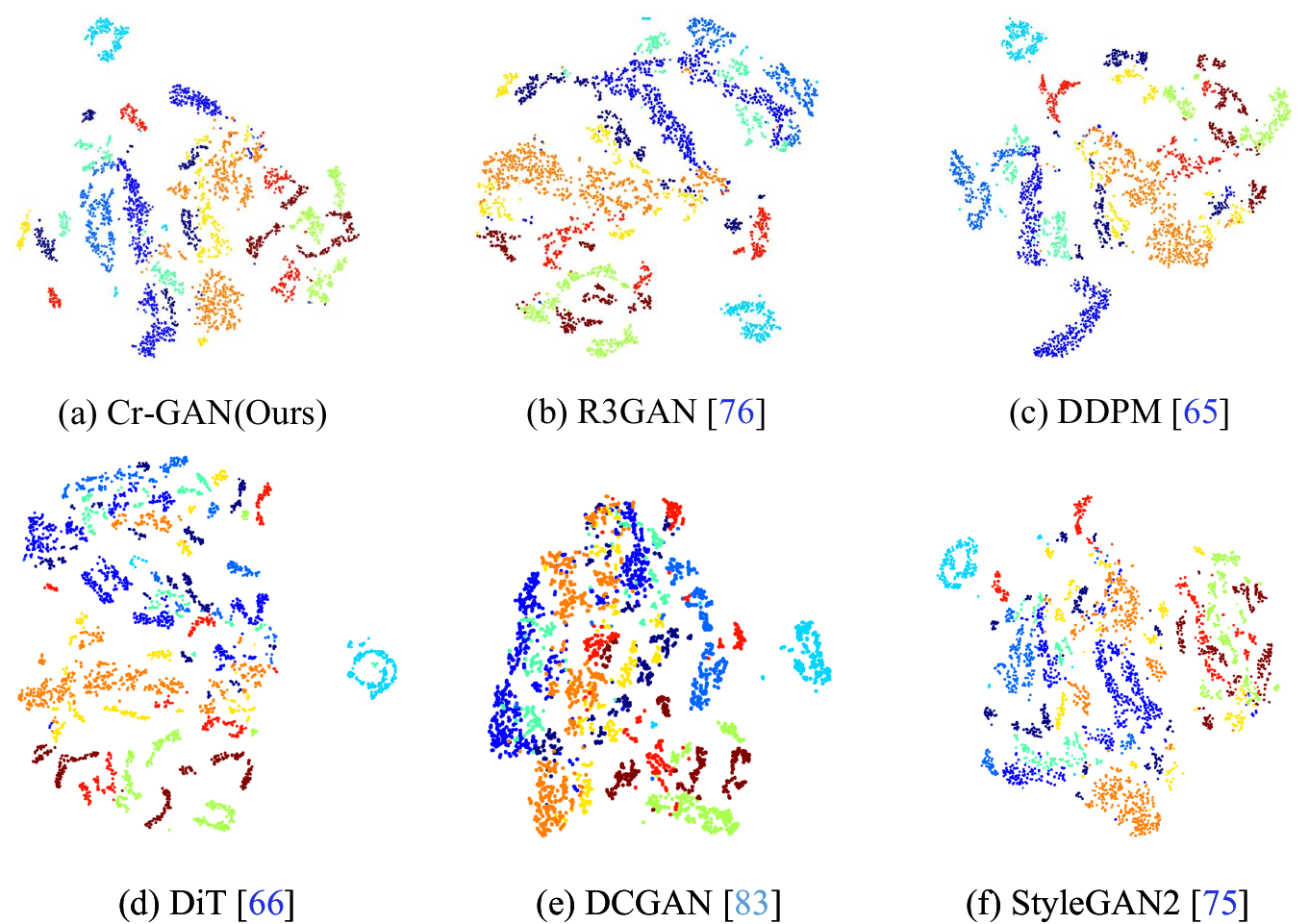}
	\caption{t-SNE visualization of the pre-trained feature space, features are from the real test set, extracted by the ResNet-18 backbone pre-trained on synthetic data from various 2-shot generative models. Cr-GAN, along with R3GAN\cite{ref45} and DDPM\cite{ref43}, successfully learn a feature space with class separability, providing a foundation for fine-tuning. The other methods (d-f) fail to learn a useful representation.}
	\label{tsne}
\end{figure}

\subsection{Visualization of Feature Representations}
To visually demonstrate that our generated images provide meaningful features for pre-training, we visualized the learned feature space. The analysis focuses on a ResNet-18~\cite{resnet} backbone pre-trained via SimCLR~\cite{ref1} on synthetic images generated by various models trained under the 2-shot condition. We passed the real test set through each pre-trained backbone to extract feature vectors, which were then projected into 2D using t-SNE~\cite{tsne}.

First, to elucidate the dynamic optimization process, Fig.~\ref{fig:tsne_dynamics} illustrates the evolution of the feature space learned by Cr-GAN across three distinct training stages.  As shown, the training initiates from a chaotic Random Initialization, where features are entangled without discernible structure. As the model is pretrained on diverse synthetic data (at Epoch 50), class-specific clusters begin to nucleate. By Epoch 100, the manifold stabilizes into compact, well-separated clusters. This trajectory visually confirms that our synthesized data actively guides the encoder to untangle the feature space, progressively learning robust decision boundaries.

Building on this robust convergence, we further compare the final learned representation of Cr-GAN against other generative baselines in Fig.~\ref{tsne}. The feature space learned from our Cr-GAN (a) is exceptionally well-structured, exhibiting clear clusters for different classes. This visualization provides strong qualitative evidence that our method produces a synthetic dataset of high utility, enabling the backbone to learn a highly discriminative representation. While other strong methods like R3GAN~\cite{ref45} (b) and DDPM~\cite{ref43} (c) show some degree of clustering, their feature spaces are less defined than ours. This suggests that while their generated data provides some benefit, it leads to a more ambiguous feature representation. The performance of the remaining baselines is far weaker. The feature spaces from DiT~\cite{ref44}, DCGAN~\cite{ref24}, and StyleGAN2~\cite{ref27} (d-f) are largely chaotic and lack any discernible class structure. Their failure to learn a useful representation from the 2-shot synthetic data directly explains their poor performance in the downstream fine-tuning tasks. This qualitative comparison highlights the superior ability of our generative process to create a diverse and semantically rich dataset, which is critical for effective pre-training in data-scarce scenarios.
\begin{table}[!t]
	\centering
	\setlength{\tabcolsep}{3pt} 
	\caption{Few-shot classification accuracy (\%) comparison on the SRSDD dataset. The best result in each setting is marked in \textbf{bold}, and the second-best result is \underline{underlined}. The improvement of our method over the second-best is shown in parentheses. ``Fail'' indicates that the model collapsed during training and produced non-informative samples (e.g., pure noise or indistinct artifacts).  The results are averaged over 3 random seeds.}
	\label{tab:srsdd}
	
	\begin{tabularx}{\columnwidth}{>{\RaggedRight}X c c c} 
		\toprule
		\textbf{Method} & \textbf{2-shot} & \textbf{4-shot} & \textbf{8-shot} \\
		\midrule
		\multicolumn{4}{l}{\textit{Diffusion-based Generative Methods}} \\
		DDPM~\cite{ref43}       & 32.53  & 37.39  & 44.77\\ 
		DiT(B/2)~\cite{ref44}   & Fail     & Fail   & Fail \\ 
		DiT(S/2)~\cite{ref44} 	& Fail     & Fail   & Fail\\
		SiT(B/2)~\cite{SiT}     & Fail   & Fail  & Fail\\ 
		SiT(S/2)~\cite{SiT}     & Fail   & Fail  & Fail\\ 
		EDM2~\cite{edm2}        & \underline{32.91}  & 38.55  & 43.91\\
		EDM2(With Pretrained)~\cite{edm2} & 32.75 & \underline{43.77} & 44.78\\
		\midrule
		\multicolumn{4}{l}{\textit{GAN-based Generative Methods}}\\
		R3GAN~\cite{ref45}      & 26.10  & 39.27 & 40.55             \\ 
		StyleGAN2 + DiffAugment~\cite{ref31} & 32.90 & 37.93 & 40.59             \\
		StyleGAN2 + DA~\cite{ref31} + $R_{LC}$~\cite{ref15} & 30.36  &  38.84  & 42.46  \\
		StyleGAN2 + ADA~\cite{ref33} & 30.58  & 41.16 & \underline{45.80}  \\ 
		DCGAN~\cite{ref24}     & 21.13   & 23.32  & 32.24  \\ 
		\rowcolor{light-gray} 
		Cr-GAN(ours) & \textbf{35.52}{\tiny (+2.62)} & \textbf{45.23}{\tiny (+1.46)} & \textbf{51.64}{\tiny (+5.84)} \\
		\bottomrule
	\end{tabularx}
\end{table}

\subsection{Analysis of Model Complexity}
To provide context for the performance results, we analyzed the computational complexity of various generative models, with key metrics presented in Table~\ref{tab:complexity}. We first note the significant scale and computational demands of modern Diffusion Models. The largest architecture in our comparison, EDM2~\cite{edm2}, utilizes nearly 280M parameters, while other prominent examples like DiT~\cite{ref44}, SiT~\cite{SiT}, and DDPM~\cite{ref43} are also characterized by large parameter counts and substantial resource requirements. Beyond their model scale, a fundamental characteristic of diffusion models is their slow, iterative sampling process, which typically requires hundreds of sequential denoising steps to generate a single image. This inherent latency can limit their practicality in applications requiring rapid synthesis.

Among the GAN-based methods, R3GAN~\cite{ref45} is notable for its exceptionally long training time (6.55 hours), a direct consequence of its gradient penalty loss, which requires an additional backpropagation pass per iteration. Similarly, while StyleGAN2~\cite{ref27} has a more moderate training time, its overall efficiency is reduced by the computationally expensive augmentation procedure embedded directly in its training loop.

In stark contrast, our proposed Cr-GAN demonstrates remarkable efficiency. It operates with only 13.71M parameters and 1.34~GFLOPs, and required just 0.27 hours of training time. This is comparable to the simplest baseline, DCGAN~\cite{ref24}, and is over 24$\times$ faster to train than R3GAN. This quantitative comparison highlights that Cr-GAN not only achieves superior downstream performance but does so with a dramatically more efficient architecture, offering a favorable balance of accuracy, training speed, and fast single-pass inference that is ideal for practical, low-resource SAR applications.
\begin{table}[!t]
	\centering
	\caption{Comparison of Model Complexity. We report the total number of trainable parameters (in Millions) and the computational cost (FLOPs(G)) for a single forward pass.}
	\label{tab:complexity}
	\resizebox{\columnwidth}{!}{
		\begin{tabular}{l c c c} 
			\toprule
			\textbf{Method} & \textbf{Parameters (M)} & \textbf{FLOPs(G)} & \textbf{Wall-Clock Time(Hours)} \\
			\midrule
			\multicolumn{3}{l}{\textit{Diffusion Models}} \\
			DDPM\cite{ref43} & 82.95 & 29.97 & 0.79\\
			DiT(B/2)\cite{ref44} & 129.56 & 87.06 & 1.16 \\
			DiT(S/2)\cite{ref44} &32.49 & 21.78& 0.41\\
			SiT(B/2)\cite{SiT} &129.56 & 87.06 & 1.25 \\
			SiT(S/2)\cite{SiT} &32.49 &21.78 & 0.43\\
			EDM2\cite{edm2} & 279.43 &  101.3 & 1.12\\
			\midrule
			\multicolumn{3}{l}{\textit{GAN Models}} \\
			DCGAN\cite{ref24} & 6.33 & 1.04 & 0.25\\
			R3GAN\cite{ref45} & 22.42 & 26.12 & 6.55\\
			StyleGAN2\cite{ref27} & 22.22 & 6.71 & 1.20\\
			\rowcolor{light-gray}Cr-GAN(ours) & 13.71 & 1.34 & 0.27 \\
			\bottomrule
		\end{tabular}
	}
\end{table}
\subsection{Ablation Study}

\textbf{Component-wise Contribution.}
To dissect the contribution of each component, we conduct a comprehensive ablation study, with results presented in Table~\ref{tab:ablation-15000}. The baseline DCGAN, without our proposed regularizations, struggles with data scarcity and yields limited performance.

Introducing our feature-cycle consistency ($\mathcal{L}_{FR}^{D}$) mechanism alone brings a significant performance gain. Its dual function of creating novel samples via feature interpolation while ensuring semantic fidelity through a cycle-consistency loss provides a strong foundation of controlled diversity. Similarly, adding only the image-space diversity loss ($\mathcal{L}_{MS}^{G}$) also improves accuracy by explicitly encouraging visually distinct outputs and thus mitigating mode collapse.

\begin{figure}[!t]
	\centering
	\includegraphics[width=0.8\columnwidth]{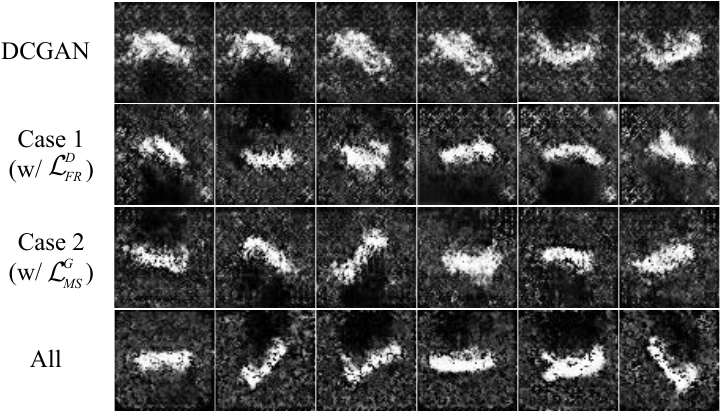}
	\caption{Qualitative ablation study demonstrating the synergistic effect of the proposed losses .
		DCGAN: Suffers from severe mode collapse with repetitive and blurry outputs.
		Case 1 (w/ $\mathcal{L}_{FR}^{D}$): Incorporating feature consistency ensures semantic fidelity, producing structurally coherent targets, though visual diversity is not fully maximized.
		Case 2 (w/ $\mathcal{L}_{MS}^{G}$) effectively encourages visual variation, but without the feature-level constraint, the generated targets occasionally lack semantic sharpness or structural coherence. 
		All: The full strategy combines both benefits, translating the interpolated latent features into high-fidelity samples with rich semantic and visual diversity.}
	\label{fig:ablation_visual}
\end{figure}

Crucially, we qualitatively verify the contribution of each component in Fig.~\ref{fig:ablation_visual}. The combination of both strategies produces the best results, revealing a powerful synergy. The  feature-cycle consistency loss  is responsible for generating diverse and meaningful content. The image-space diversity loss then acts as a complementary regularizer, guaranteeing this latent diversity is fully realized with high visual variety in the final images. This comprehensive approach to enhancing both fidelity and diversity directly leads to the superior downstream classification performance.

\begin{table}[! t]
\centering
\caption{Component Ablation study. \ding{51} denotes the presence of a component.}
\label{tab:ablation-15000}
\setlength{\tabcolsep}{3pt}
\begin{tabular}{lccc|ccc}
\toprule
\textbf{Configuration} & $\mathcal{L}^{D}_{FR}$ & $\mathcal{L}^{G}_{MS}$ & & \textbf{2-shot} & \textbf{4-shot} & \textbf{8-shot} \\
\midrule
Baseline(DCGAN\cite{ref24}) & & & & 36.49 & 51.63 & 68.50 \\
Case1 & \ding{51} & & & 48.37 & 54.04 & 69.37 \\
Case2 & & \ding{51} & & 45.76 & 53.01 & 71.20 \\
\rowcolor{light-gray}All & \ding{51} & \ding{51} & & \textbf{47.24}& \textbf{60.06} & \textbf{71.21} \\
\bottomrule
\end{tabular}
\end{table}

\begin{table}[! t]
	\centering
	\caption{Effect of training iterations on downstream performance.}
	\label{tab:checkpoints}
	\begin{tabularx}{\columnwidth}{>{\RaggedRight}X >{\centering}X >{\centering}X >{\centering\arraybackslash}X}
		\toprule
		\textbf{Checkpoint} & \textbf{2-shot} & \textbf{4-shot} & \textbf{8-shot} \\
		\midrule
		5000  & 43.74 & 57.70 & 69.80 \\
		10000 & 47.23 & 58.00 & 64.06 \\
		\rowcolor{light-gray}15000 & \textbf{47.24} & \textbf{60.06} & \textbf{71.21} \\
		20000 & 47.21 & 60.03 & 71.10\\
		\bottomrule
	\end{tabularx}
\end{table}

\begin{figure}[t]
	\centering
	\includegraphics[width=0.7\columnwidth]{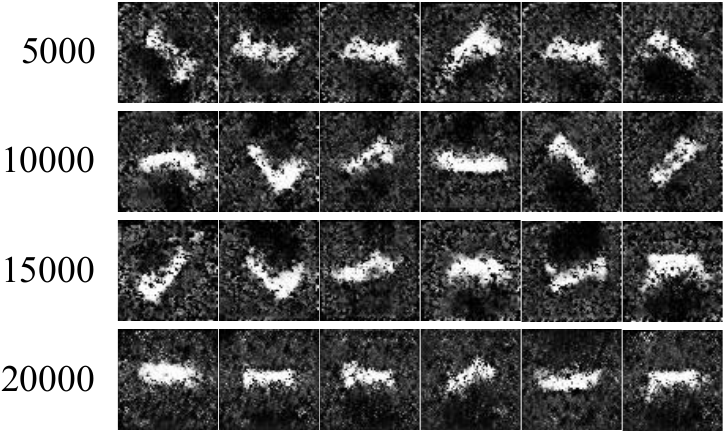}
	\caption{Visual evolution of synthesized samples across training iterations (corresponding to Table VI). 
		5000-10000 iter: Generated images lack definition, with blurred edges and weak scattering features.
		15000 iter: The generator achieves optimal performance, producing targets with sharp details and realistic backgrounds.
		20000 iter: Visual quality saturates, showing no significant improvement over the 15k checkpoint.}
	\label{V}
\end{figure}

\textbf{Effect of Training Iterations.} 
To determine the optimal training duration for our generative model, we conducted an ablation study on its training checkpoints, with results summarized in Table~\ref{tab:checkpoints}. We evaluated three generator checkpoints at 5k, 10k, 15k, 20k iterations, the downstream performance in the 2-shot and 4-shot scenarios shows a clear positive correlation with training time, improving from 43.74\% to 47.24\% and 57.70\% to 60.06\%, respectively. The 8-shot setting exhibits a different, non-monotonic pattern, with performance decreasing from 69.80\% at 5k to 64.06\% at 10k, before reaching its overall peak of 71.21\% at the 15k checkpoint. Despite this fluctuation, the 15,000-iteration checkpoint consistently achieves the highest performance across all three few-shot settings. To complement the quantitative results in Table~VI, Fig.~\ref{V} illustrates the quality evolution of generated samples. Therefore, to ensure the best possible foundation for our main experiments, this 15k checkpoint was selected.
\begin{figure}[!t]
	\centering
	\includegraphics[width=\columnwidth]{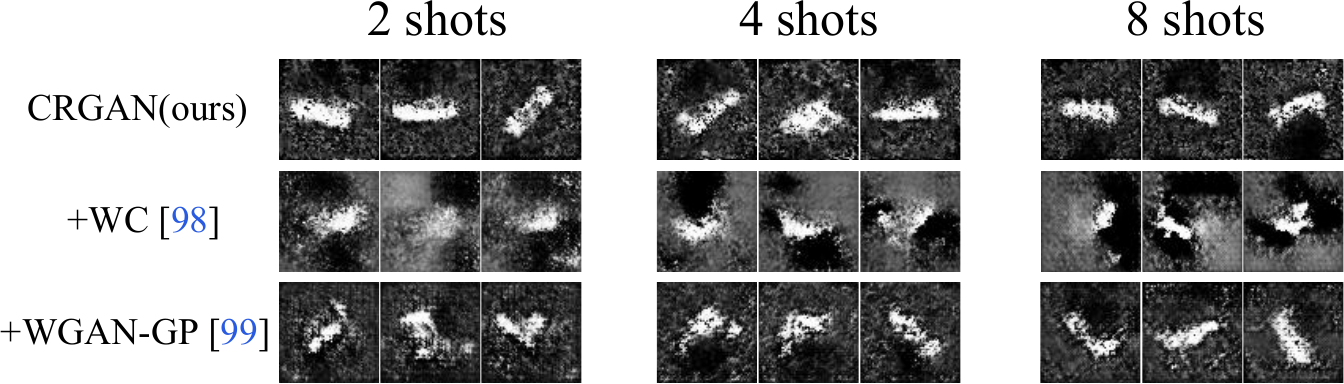}
	\caption{Visual comparison of different GAN stabilization techniques across 2, 4, and 8-shot settings. 
		Cr-GAN (ours): Demonstrates consistent stability and high target fidelity even with extremely limited data.
		+WC \cite{wc}: Introducing Weight Clipping leads to severe quality degradation, resulting in blurry and noisy outputs.
		+WGAN-GP \cite{wgan}: While Gradient Penalty produces reasonable images at 8 shots, it offers no distinct visual advantage over our method in the more challenging 2-shot and 4-shot regimes.}
	\label{fig:ablation_stabilization}
\end{figure}

\begin{table}[!t] 
	\centering
	\caption{Ablation study on GAN stabilization techniques. We report downstream classification accuracy (\%) on the MSTAR dataset.The $+$ symbol denotes the addition of a given technique to our method.}
	\label{tab:gan_stabilization}
	\begin{tabularx}{\columnwidth}{>{\RaggedRight}X c c c}
		\toprule
		\textbf{Stabilization Method} & \textbf{2-shot} & \textbf{4-shot} & \textbf{8-shot} \\
		\midrule
		\rowcolor{light-gray} CRGAN(ours) & \textbf{47.24} & \textbf{60.06} & 71.21 \\
		+ WC \cite{wc} & 35.37 & 54.29 & 66.96 \\
		+ WGAN-GP \cite{wgan} & 39.90 & 59.16 & \textbf{72.52} \\ 
		\bottomrule
	\end{tabularx}
\end{table}

\textbf{GAN Stabilization Techniques}. To validate our approach to training stability, we conducted an ablation study comparing different GAN stabilization techniques, with results presented in Table~\ref{tab:gan_stabilization}. Our primary model, denoted as CRGAN (ours), relies on our proposed consistency regularizations and the Spectral Normalization\cite{SN} already integrated into our modern discriminator architecture, without additional explicit stabilizers. We compare this against two variants: one incorporating the outdated Weight Clipping (+ WC)\cite{wc}, and another enhanced with the modern Gradient Penalty (+ WGAN-GP)\cite{wgan}.
The results demonstrate that utilizing our proposed CR as the sole regularizer yields highly competitive performance, validating its sufficiency and effectiveness for stabilizing training in the extreme few-shot regime. While the integration of a Gradient Penalty\cite{wgan} offers a synergistic improvement in the 8-shot setting to a peak of 72.52\%, its performance is inferior to using our CR method alone in the more extreme 2-shot and 4-shot scenarios. Besides, applying Weight Clipping\cite{wc} severely degrades performance, confirming its unsuitability. Therefore, we adopt it as the sole primary regularization strategy for the discriminator in all main experimental settings.\par

As visualized in Fig.~\ref{fig:ablation_stabilization}, applying Weight Clipping (+WC)\cite{wc} leads to severe blurriness and degradation. While Gradient Penalty (+WGAN-GP)\cite{wgan} produces reasonable results, it offers no distinct visual advantage over our method in the challenging 2-shot setting. This confirms that our consistency regularization is robust enough to stabilize training without additional constraints.

\begin{figure*}[! t] 
	\centering
	\includegraphics[width=\textwidth]{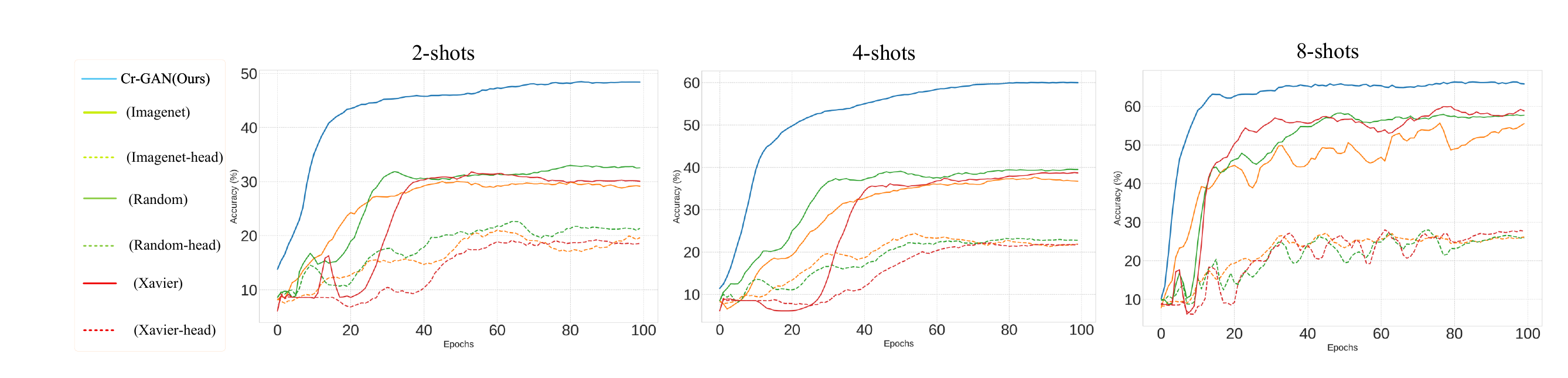}
	\caption{Performance comparison of our method against various fine-tuning baselines under 2-shot, 4-shot, and 8-shot settings. The baselines include ImageNet pre-training, Xavier initialization, and random initialization, with fine-tuning applied to either the full network (solid lines) or only the classification head (dashed lines). Our method demonstrates superior performance across all configurations.}
	\label{fig:ablation_finetune}
\end{figure*}

\textbf{Comparison with Fine-tuning under Various Initialization.}
To comprehensively validate the effectiveness of our proposed method, we conduct a series of ablation studies, comparing it against several traditional fine-tuning approaches based on different initialization strategies. Specifically, we consider three baseline initialization schemes: (1) using weights pre-trained on ImageNet\cite{ref35}, (2) using Xavier\cite{xavier} initialization, and (3) using standard random initialization. For each scheme, we evaluate two fine-tuning strategies: one that fine-tunes the entire network (full fine-tuning) and another that only fine-tunes the final classification head (denoted by the ``-head'' suffix), while keeping the backbone network frozen. The comparisons are conducted under three typical few-shot learning scenarios: 2-shots, 4-shots, and 8-shots, with the results presented in Fig~\ref{fig:ablation_finetune}.
\FloatBarrier

\begin{figure}[! t] 
	\centering
	\includegraphics[width=\columnwidth]{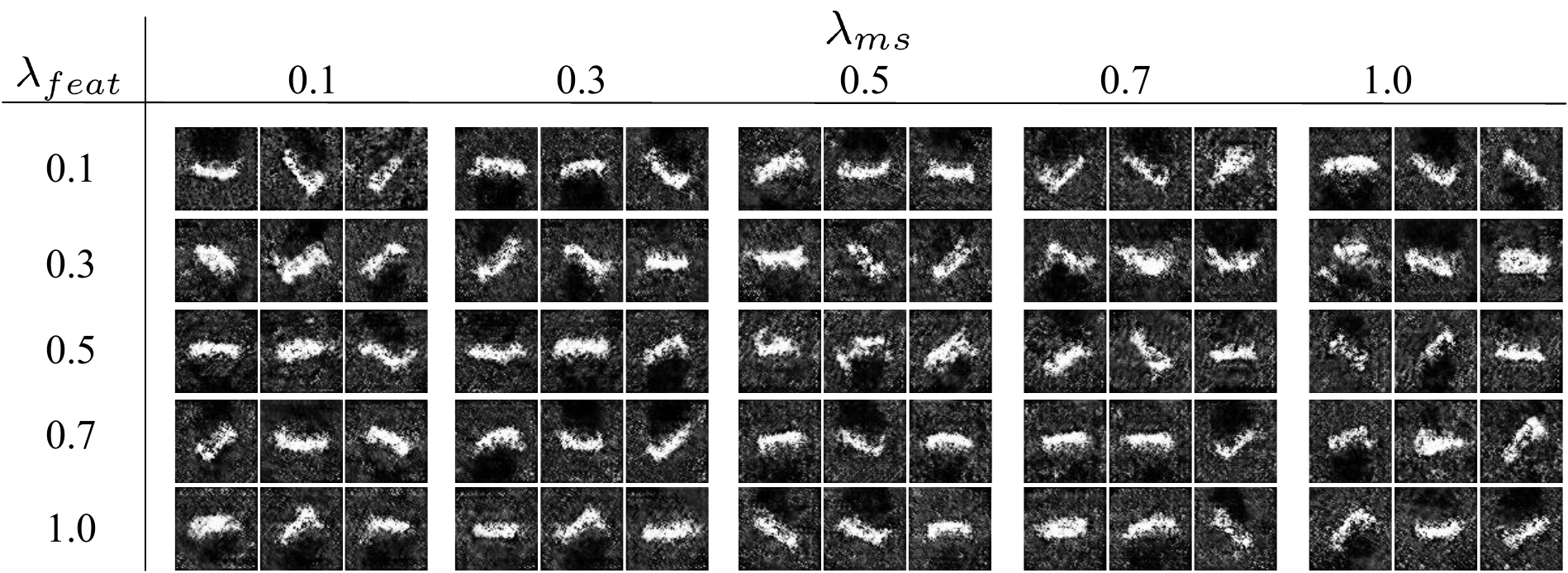}
	\caption{Visual ablation study on the interaction between $\lambda_{feat}$ weight and $\lambda_{ms}$ weight.}
	\label{VII}
\end{figure}

\begin{table}[! t]
	\centering
	\caption{Ablation study on the loss weights $\lambda_{feat}$ and $\lambda_{ms}$, with accuracy (\%) on the validation set reported.}
	\label{tab:hyperparameter_ablation}
	\begin{tabular}{lccccc}
		\toprule
		& \multicolumn{5}{c}{\textbf{$\lambda_{ms}$}} \\
		\cmidrule(lr){2-6}
		\textbf{$\lambda_{feat}$} & 0.1 & 0.3 & 0.5 & 0.7 & 1.0 \\
		\midrule
		0.1 & \colorbox{light-gray}{\textbf{47.24}} & 47.16 & 47.04 & 44.24 & 45.52 \\
		0.3 & 44.86 & 43.80 & 47.08 & 46.02 & 44.40 \\
		0.5 & 46.61 & 46.21 & 41.90 & 42.90 & 44.21 \\
		0.7 & 46.77 & 46.74 & 42.93 & 43.02 & 43.43 \\
		1.0 & 45.02 & 43.77 & 44.71 & 44.52 & 47.02 \\
		\bottomrule
	\end{tabular}
\end{table}

\textbf{Ablation Study on Loss Weights.}
We performed an ablation study to investigate the impact of the  ($\lambda_{feat}$) and  ($\lambda_{ms}$) on model's 2-shot performance. The empirical results, presented in Table~\ref{tab:hyperparameter_ablation}, reveal a clear performance trend. We observed that lower weight values for both loss components consistently yielded higher accuracy, with the peak performance of 47.24\% achieved at $\lambda_{feat}=0.1$ and $\lambda_{ms}=0.1$. To visually verify the impact of these hyperparameters, Fig.~\ref{VII} presents a grid of generated samples. Excessively high $\lambda_{feat}$ enforces alignment with the interpolated vector, the strong uniformity constraint forcibly pushes the generated features away from the real data distribution. This conflict with the adversarial objective drives the latent codes into low-density regions of the manifold, resulting in severe high-frequency noise and artifacts as the generator struggles to reconcile the divergence, while overly strong $\lambda_{ms}$ introduces excessive high-frequency noise in an attempt to force pixel-level diversity. This visual evidence corroborates the quantitative findings, confirming that a moderate regularization strength yields the optimal balance between diversity and fidelity. This suggests that a more conservative weighting is beneficial for our task. Based on this observation, we selected $\lambda_{feat}=0.1$ and $\lambda_{ms}=0.1$ for our main experiments.\par
\begin{figure}[! t]
	\centering
	\includegraphics[width=0.6\columnwidth]{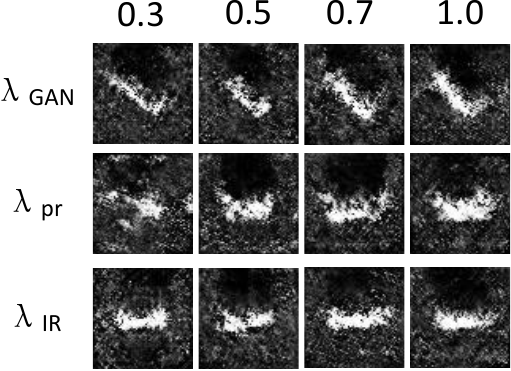}
	\caption{Visual ablation study on the sensitivity of loss weights ($\lambda_{GAN}$, $\lambda_{pr}$, and $\lambda_{IR}$). Each row displays generated samples when varying one specific weight from 0.3 to 1.0 while keeping others 1.0. }
	\label{VIII}
\end{figure}

Furthermore, we validated our choice for the weights of the foundational loss components: the adversarial loss ($\lambda_{GAN}$), image reconstruction loss ($\lambda_{IR}$), and prior regularization loss ($\lambda_{pr}$). In the interest of maintaining a fair comparison with baseline VAE-GAN frameworks, we set their default values to 1.0. To empirically verify this choice, we conducted a sensitivity analysis on the 2-shot MSTAR task, varying each weight individually while holding the others constant. As presented in Table~\ref{tab:weight_ablation}, the results unequivocally support our initial setting.To provide an intuitive understanding of these hyperparameters, we visualize the generated samples under different weight configurations in Fig.~\ref{VIII}. As observed, reducing $\lambda_{GAN}$ diminishes the ability to recover high-frequency details, resulting in blurred targets and loss of speckle texture. A low $\lambda_{pr}$ fails to constrain the latent posterior, leading to severe artifacts and background noise due to sampling from a fragmented manifold. Similarly, insufficient $\lambda_{IR}$ weakens semantic anchoring, causing structural deformation where the generator fails to preserve the target's geometry. Consequently, the model's performance consistently degrades when any of the three core weights are reduced from 1.0, with the peak accuracy achieved when $\lambda_{GAN} = \lambda_{IR} = \lambda_{pr} = 1.0$.\par
\begin{table}[!t]
	\centering
	\caption{Analysis for  loss weights on the 2-shot, reporting classification accuracy (\%).}
	\label{tab:weight_ablation}
	\begin{tabular}{cccc}
		\toprule
		\textbf{$\lambda_{\mathrm{GAN}}$} & \textbf{ $\lambda_{\mathrm{pr}}$} & \textbf{$\lambda_{\mathrm{IR}}$} & \textbf{Accuracy (\%)} \\
		\midrule
		0.3 & 1.0   & 1.0   & 43.05 \\
		0.5 & 1.0   & 1.0   & 44.27 \\
		0.7 & 1.0   & 1.0   & 46.58 \\
		\midrule
		1.0   & 0.3 & 1.0   & 42.77 \\
		1.0   & 0.5 & 1.0   & 43.21 \\
		1.0   & 0.7 & 1.0   & 44.55 \\
		\midrule
		1.0   & 1.0   & 0.3 & 40.34 \\
		1.0   & 1.0   & 0.5 & 42.49 \\
		1.0   & 1.0   & 0.7 & 44.68 \\
		\midrule
		\rowcolor{light-gray} \textbf{1.0} & \textbf{1.0} & \textbf{1.0} & \textbf{47.24} \\
		\bottomrule
	\end{tabular}
\end{table}
\vspace{0pt}
\textbf{Comparison With Self-supervised Learning.}
To validate our choice of SimCLR\cite{ref1} for the pre-training stage, we conducted an ablation study comparing it against other prominent SSL  methods. We used the same synthetic dataset, generated by our 15k generator checkpoint, to pre-train a ResNet-18 backbone with four different SSL methods: SimCLR, BiDFC\cite{WCL}, MoCo\cite{ref2}, and BYOL\cite{ref3}. The effectiveness of each pre-trained backbone was then evaluated on the downstream few-shot classification tasks.
The results, summarized in Table~\ref{tab:ssl_ablation}, clearly demonstrate the effectiveness of our chosen approach.
\begin{table}[!t]
\centering
\caption{Performance of different SSL methods on few-shot tasks. All values represent downstream classification accuracy in percent (\%).}
\label{tab:ssl_ablation}
\begin{tabularx}{\columnwidth}{>{\RaggedRight}X >{\centering}X >{\centering}X >{\centering\arraybackslash}X}
\toprule
\textbf{SSL Method} & {\textbf{2-shot}} & {\textbf{4-shot}} & {\textbf{8-shot}}  \\
\midrule
BiDFC\cite{WCL} & 46.51 & 59.13 & 70.45  \\
MoCo\cite{ref2} & 45.33 & 57.82 & 68.91  \\
BYOL\cite{ref3} & 44.19 & 57.05 & 67.68  \\
\rowcolor{light-gray}SimCLR\cite{ref1} & \bfseries 47.24 & \bfseries 60.06 & \bfseries 71.21  \\
\bottomrule
\end{tabularx}
\end{table}

\textbf{Comparison of GAN Architectures.}
To assess the generalizability of our framework, we conducted a comparative study by incorporating its core components (i.e., the dual-branch discriminator and consistency regularization) into two other prominent generator architectures: StyleGAN2\cite{ref27} and BigGAN\cite{ref25}. All training settings, including feature interpolation and SimCLR pre-training, remained constant. The results in Table~\ref{tab:gan_ablation} demonstrate that our complete model, utilizing the DCGAN-based architecture, achieves the highest classification accuracy across all few-shot settings. Interestingly, the more complex StyleGAN2\cite{ref27} and BigGAN\cite{ref25} architectures, while powerful on large datasets, yielded lower performance, suggesting their higher parameter counts may predispose them to overfitting in data-scarce SAR scenarios. This study reveals a crucial insight: for tasks defined by extreme data scarcity, the effectiveness of the training and regularization strategy can be more critical than the raw complexity of the generator architecture.
\begin{table}[! t]
	\centering
	\caption{Comparison of Our Consistency-Regularized Framework on Different GAN Architectures.}
	\label{tab:gan_ablation}
	\begin{tabularx}{\columnwidth}{>{\RaggedRight}X >{\centering}X >{\centering}X >{\centering\arraybackslash}X}
		\toprule
		\textbf{Architecture} & {\textbf{2-shot}} & {\textbf{4-shot}} & {\textbf{8-shot}}  \\
		\midrule
		StyleGAN2~\cite{ref27}  & 45.37 & 56.78 & 68.01\\
		BigGAN~\cite{ref25}     & 46.21 & 55.03 & 66.38 \\
		\rowcolor{light-gray}
		DCGAN~\cite{ref24}      & \textbf{47.24} & \textbf{60.06} & \textbf{71.21} \\
		\bottomrule
	\end{tabularx}
\end{table}
\subsection{About Fréchet Inception Distance}
Evaluating the quality of generated images is a critical yet challenging aspect of our work. While Fréchet Inception Distance (FID)\cite{fid} is a widely adopted metric for natural image generation, its applicability to the RS domain is questionable. FID relies on an InceptionV3\cite{inceptionv3} model pre-trained on ImageNet\cite{ref35}, which creates a significant domain gap. The feature space of InceptionV3 is optimized for classifying everyday objects and scenes, and is thus ill-suited to capture the unique top-down perspectives, complex textures, and specific spectral properties inherent in SAR imagery.
Consequently, FID\cite{fid} scores in this context may not reliably correlate with perceptual quality or the utility of the generated data. An evaluation metric should ideally reflect the model's ability to produce useful samples for a given purpose. Therefore, instead of relying on a potentially misleading FID score, we argue that a more pragmatic and domain-relevant evaluation is to measure the impact of the generated data on a downstream task.
For this reason, we adopt the classification accuracy on a few-shot recognition task as our primary metric for assessing generation quality. This approach directly quantifies the practical value of the synthetic images by evaluating how well they can augment a training set to improve model performance. This downstream-oriented evaluation provides a more meaningful and reliable assessment of our model's capabilities in the RS context.

\section{Conclusion}
This paper proposes Cr-GAN to address the problem of SAR target recognition in the context of labeled data scarcity. The proposed model integrates a dual-branch discriminator with feature-level interpolation and contrastive regularization, enabling the generation of diverse and semantically aligned synthetic samples. Through extensive experiments on SAR target recognition tasks, we demonstrate that Cr-GAN significantly improves few-shot classification performance compared to existing GAN and diffusion-based approaches, while maintaining a lightweight model complexity. The effectiveness of the proposed shortcuts is validated through comprehensive ablation studies. These results indicate that Cr-GAN provides a practical and efficient solution for generative modeling in low-resource RS scenarios.

Although diffusion models have recently shown remarkable generation quality, their large parameter size and slow iterative inference limit their practicality in few-shot and resource-constrained RS scenarios. In contrast, our Cr-GAN offers a superior trade-off between diversity, performance, and efficiency. A compelling direction for future work is therefore to treat our core ideas, feature-level consistency constraints and dual-branch discrimination, as a versatile regularization framework and explore their integration into other advanced generative models, including diffusion architectures. Our goal is to enhance their data efficiency and sample quality in scarce conditions, thereby providing a powerful foundation for a wider range of downstream applications, including challenging dense prediction tasks like object detection and semantic segmentation.

\bibliographystyle{IEEEtran} 
\bibliography{references}

\vspace{0pt}
\begin{IEEEbiography}[{\includegraphics[width=1in,height=1.25in,clip,keepaspectratio]{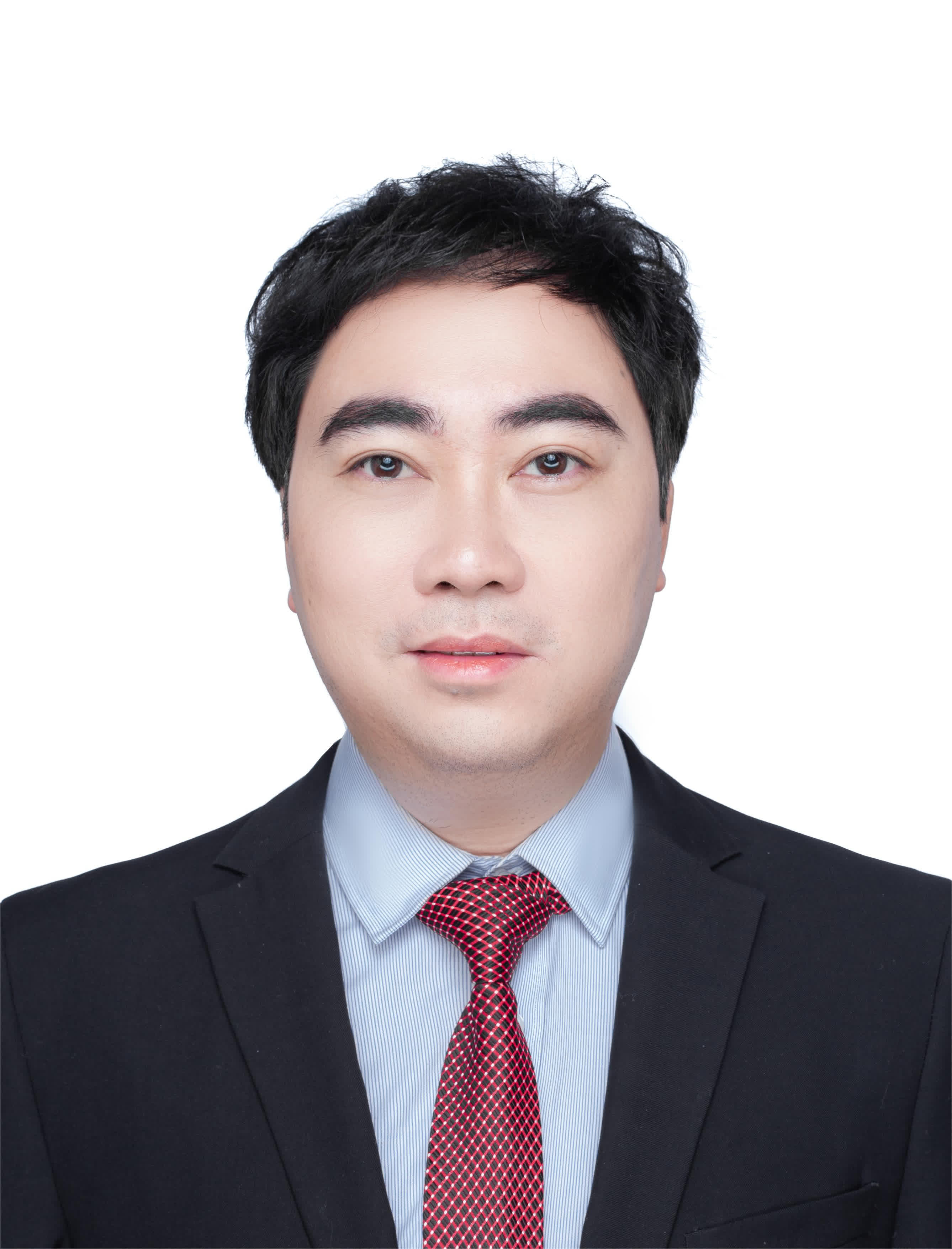}}]{Yikui Zhai}
	(Senior Member, IEEE) received his Ph.D. degree in signal and information processing from Beihang University, Beijing, China, in June 2013. Since October 2007, he has been working with Wuyi University, Jiangmen, China, where he is a Full Professor now. He is also Associate Dean of the School of Electronics and Information Engineering in Wuyi University, since 2024. He has been a Visiting Scholar with Department of Computer Science, the Universita degli Studi di Milano, Italy, during June 2016 to June 2017, August 2023 and January 2024. His research interests include: Image Processing, Deep Learning, Optical Character Recognition, Object Detection, UAV Change Detection, Self Supervise Learning.
\end{IEEEbiography}
\vspace{0pt}
\begin{IEEEbiography}[{\includegraphics[width=1in,height=1.25in,clip,keepaspectratio]{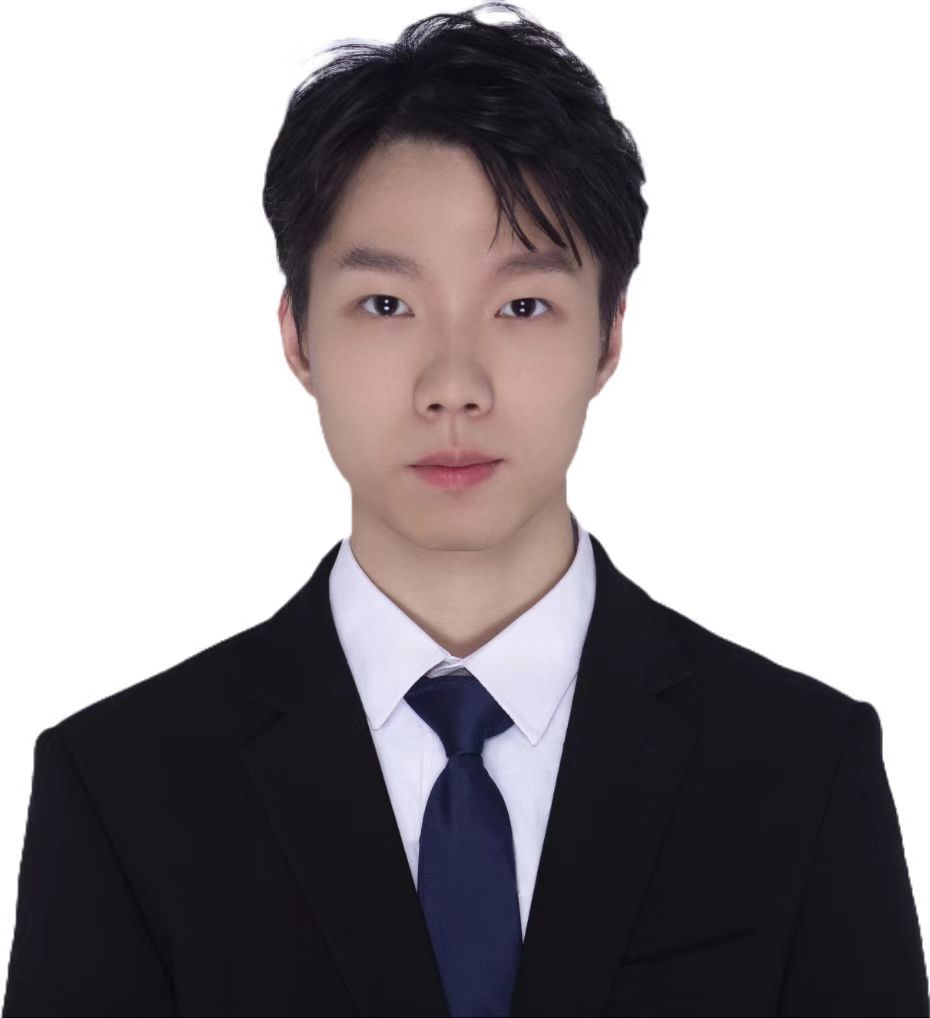}}]{Shikuang Liu}
	received the B.S. degree from Wuyi University, Jiangmen, China,
	in 2024. He is pursuing the master’s degree with the
	College of Electronics and Information Engineering,
	Wuyi University, Jiangmen, China.
	His research interests include computer vision,
	 image generation and representation learning.
\end{IEEEbiography}
\vspace{0pt}
\begin{IEEEbiography}[{\includegraphics[width=1in,height=1.25in,clip,keepaspectratio]{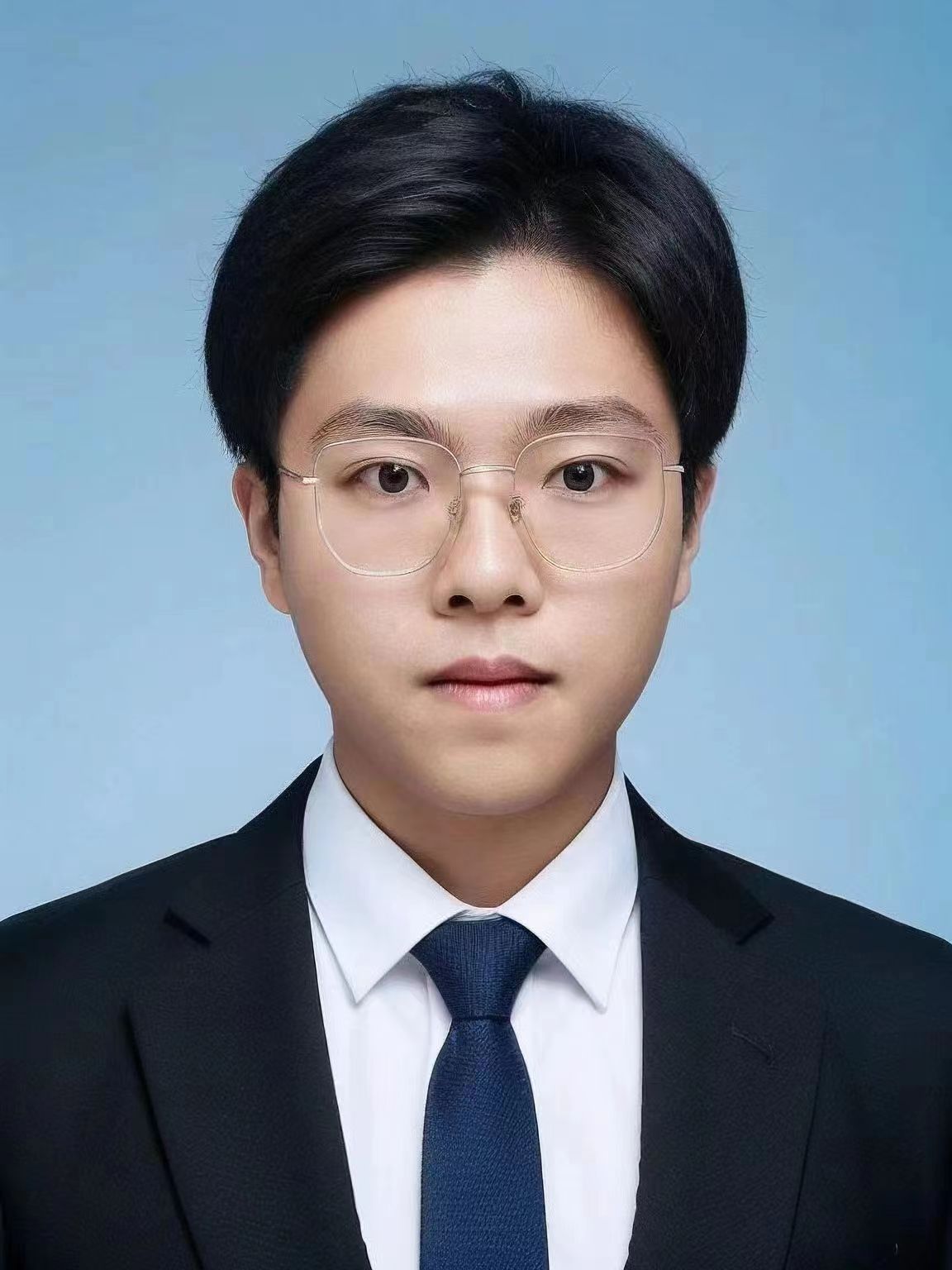}}]{Wenlve Zhou}
	received the M.S. degree in Information and Communication Engi-neering from Wuyi University, Jiangmen, China. He is currently pursuing the Ph.D. degree in Information and Communi-cation Engineering from South China University of Technology, Guangzhou, China. 
	His research interests include: computer vision, especially transfer learning and representation learning.
\end{IEEEbiography}
\vspace{0pt}

\begin{IEEEbiography}[{\includegraphics[width=1in,height=1.25in,clip,keepaspectratio]{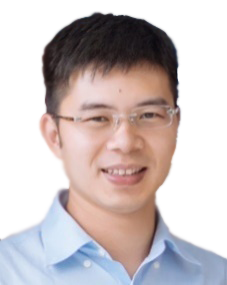}}]{Hongsheng Zhang}
	(Senior Member, IEEE) received
	the B.Eng. degree in computer science and technology and the M.Eng. degree in computer applications technology from South China Normal University,
	Guangzhou, China, in 2007 and 2010, respectively,and the Ph.D. degree in earth system and geoinformation science from The Chinese University of
	Hong Kong, Hong Kong, China, in 2013. He is currently an Assistant Professor with the Department of Geography, The University of Hong Kong, Hong Kong, China. 
	His research interests include remote-sensing applications in tropical and subtropical areas, with a focus on the urban environment and coastal sustainability monitoring, using multisource remote sensing data fusion and image pattern recognition
	techniques.
\end{IEEEbiography}
\vspace{24pt}
\begin{IEEEbiography}[{\includegraphics[width=1in,height=1.25in,clip,keepaspectratio]{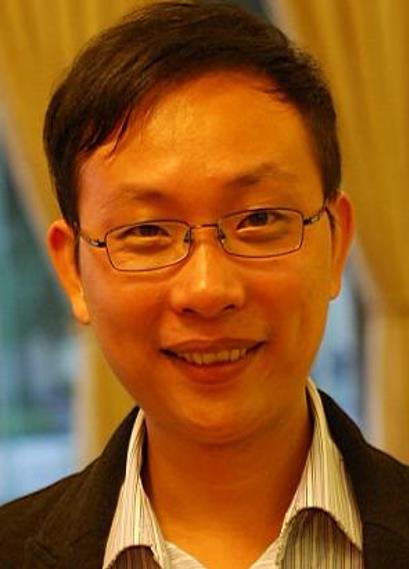}}]{Zhiheng Zhou}
	received the B.S. and M.S. degrees in Applied Mathematics and the Ph.D. degree in Electronic and Information Engineering from South China University of Technology, Guangzhou, China, in 2000, 2002, and 2005, respectively. 
	He is currently a Professor with South China University of Technology. 
	His research interests include image processing and image and video transmission.
\end{IEEEbiography}

\begin{IEEEbiography}[{\includegraphics[width=1in,height=1.25in,clip,keepaspectratio]{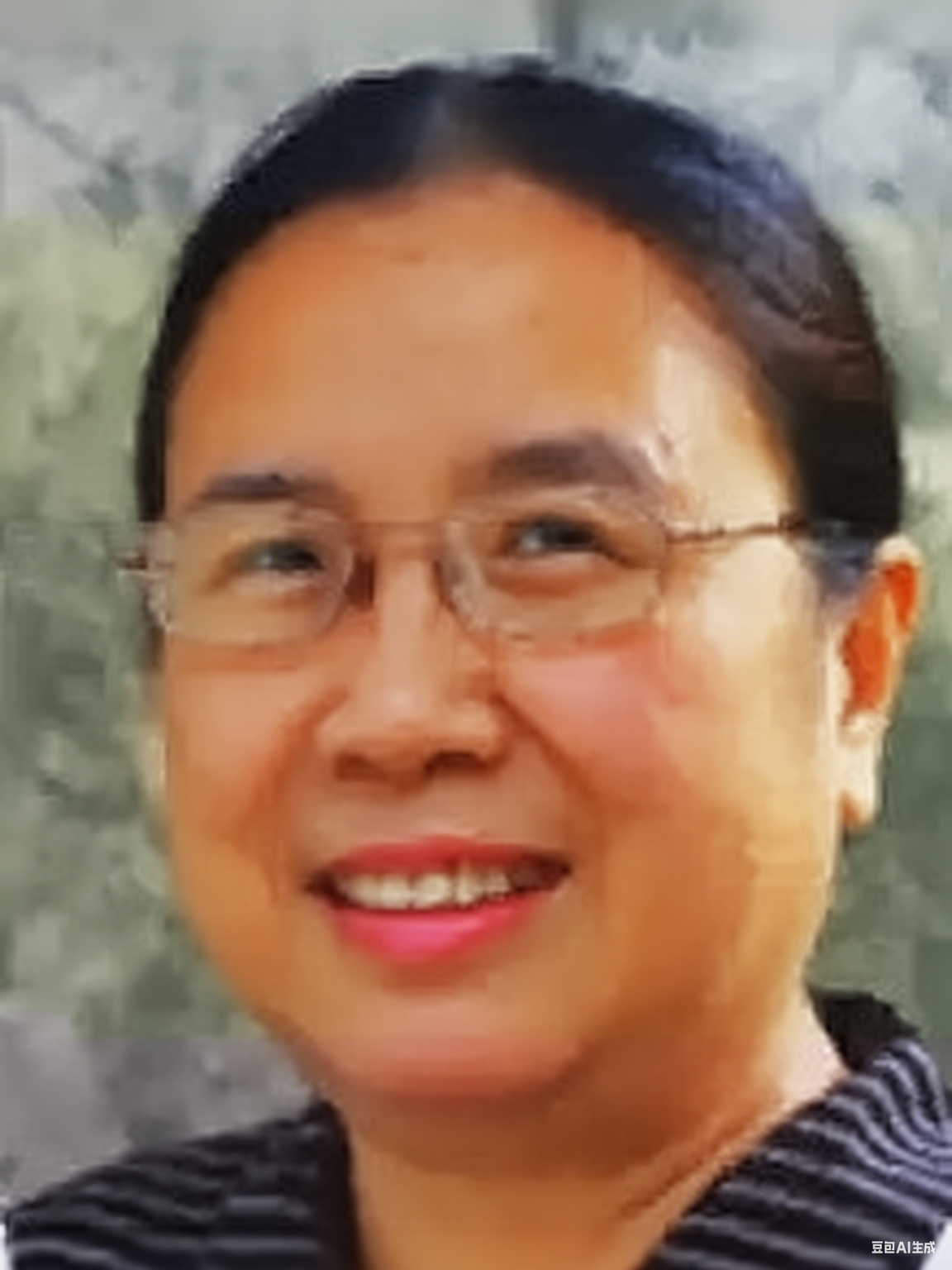}}]{Xiaolin Tian}
	graduated from Peking University, Beijing, China. She was a Lecturer with Peking University, in 1982. She was an Associate Professor with Peking University, in 1989. She was a Visiting Scholar with the Artificial Intelligence Laboratory, Computer Vision Group, Department of Electrical Engineering and Computer Science (EECS), University of Michigan, Ann Arbor, MI, USA, in 1989; and a Research Associate with the Institute for Advanced Computer Studies, University of Maryland, Collage Park, MD, USA, in 1990. She is currently a Professor with the Faculty of Information Technology and the Lunar and Planetary Science Laboratory/Space Science Institute, Macau University of Science and Technology, Macau. She is also working on medical image processing and the intelligent computing and automatic processing for huge data of change. Her research interests include image processing and pattern recognition.
\end{IEEEbiography}
\vspace{0pt}
\begin{IEEEbiography}[{\includegraphics[width=1in,height=1.25in,clip,keepaspectratio]{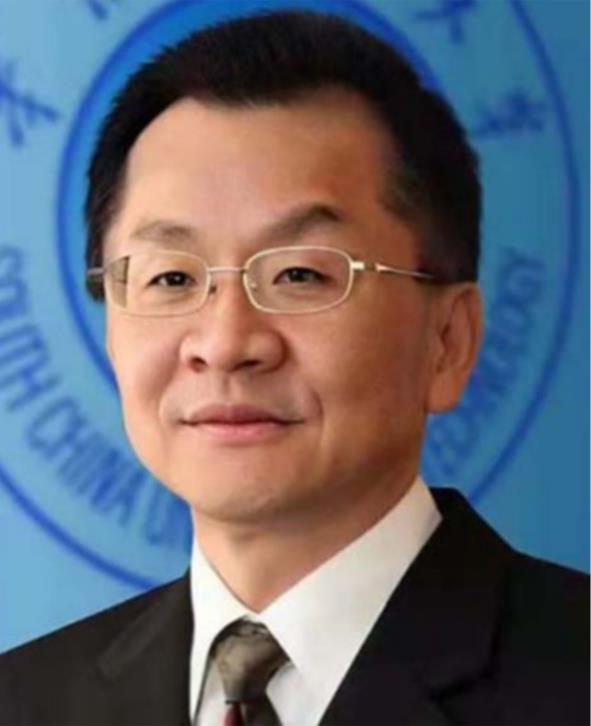}}]{C.~L.~Philip~Chen}
	(Fellow, IEEE) received the M.S. degree in electrical engineering from the University of Michigan at Ann Arbor, Ann Arbor, MI, USA, in 1985, and the Ph.D. degree in electrical engineering from Purdue University, West Lafayette, IN, USA, in 1988. 
	
	He was a tenured Professor, the Department Head, and an Associate Dean of two different universities in the U.S. for 23 years. He is currently the Head of the School of Computer Science and Engineering, South China University of Technology, Guangzhou, Guangdong, China. His current research interests include systems, cybernetics, and computational intelligence. 
	
	Dr. Chen received the 2016 Outstanding Electrical and Computer Engineers Award from his alma mater, Purdue University. He has been the Editor-in-Chief of the IEEE TRANSACTION ON SYSTEMS, MAN, AND CYBERNETICS: SYSTEMS since 2014 and an associate editor of several IEEE TRANSACTIONS. He was the Chair of TC 9.1 Economic and Business Systems of International Federation of Automatic Control from 2015 to 2017 and also a Program Evaluator of the Accreditation Board of Engineering and Technology Education of the U.S. for computer engineering, electrical engineering, and software engineering programs. He was the IEEE SMC Society President from 2012 to 2013 and a Vice President of Chinese Association of Automation (CAA). He is a Fellow of AAAS, IAPR, CAA, and HKIE.
\end{IEEEbiography}
\end{document}